\pdfoutput=1
\documentclass[10pt, a4paper]{article}

\usepackage{lrec2026}

\usepackage{amsmath}
\usepackage{booktabs}
\usepackage{siunitx}
\usepackage{multirow}
\usepackage{array}
\usepackage{caption}
\usepackage{subfigure}
\usepackage{subcaption}

\usepackage{xcolor}
\usepackage{tcolorbox}
\usepackage{ifthen}
\usepackage{titlesec}
\usepackage{multicol} 
\usepackage{adjustbox}

\newcommand{\Answer}[3]{%
  \textbf{#1: }%
  \ifthenelse{\equal{#3}{correct}}%
    {\textcolor{green!100!white}{#2}}%  % 30% green, 70% white
    {\textcolor{red!100!white}{#2}}%    % 30% red, 70% white
  \\[0.5em]%
}

\usepackage[T1]{fontenc}
\usepackage[utf8]{inputenc}

\usepackage{booktabs}   % For \toprule, \midrule, \bottomrule
\usepackage{tabularx}   % For flexible-width columns

\usepackage{microtype}

\usepackage{inconsolata}

\usepackage{graphicx}

\usepackage{tcolorbox}
\usepackage{tabularx} % For adjustable-width columns

\usepackage{xcolor}
\usepackage{tikz}
\usepackage{booktabs}  % for nicer tables
\usepackage{pgfmath}   % for arithmetic

\definecolor{DistinctYellow}{HTML}{fbec5d } % Golden Yellow
\definecolor{DistinctGreen}{HTML}{228B22}  % Lime Green
\definecolor{DistinctBlue}{HTML}{1e90ff}    % Bright Blue

\newcommand{\highlighttoken}[2]{%
  \pgfmathsetmacro{\perc}{100*#2}%
  \pgfmathparse{#2 < 0.5 ? 1 : 0}%
  \ifdim\pgfmathresult pt=1pt
    \def\colorinterp{DistinctGreen!\perc!DistinctYellow}%
  \else
    \def\colorinterp{DistinctBlue!\perc!DistinctYellow}%
  \fi
  \tikz[baseline=(token.base)]{
    \node[
      inner sep=1mm,
      anchor=base,
      fill=\colorinterp,
      rounded corners,
      font=\small
    ] (token) {#1};
  }%
}

\newcommand{\probsclhoriz}{%
  \begin{tikzpicture}[baseline=(base)]
    \coordinate (base) at (0,0);
    \shade[
      left color=DistinctYellow,
      middle color=DistinctGreen,
      right color=DistinctBlue
    ] (0,0) rectangle (6,0.4);
    \foreach \x/\val in {0/0.0,1.2/0.2,2.4/0.4,3.6/0.6,4.8/0.8,6/1.0}{
      \draw (\x,0.4) -- (\x,0.5);
      \node[above, font=\small] at (\x,0.5) {\val};
    }
  \end{tikzpicture}%
}

\newtcolorbox{llmprompt}[1][]{
  colback=gray!20, 
  colframe=gray!70,
  fonttitle=\bfseries\ttfamily,
  title=#1, % Use the provided title
  rounded corners, 
  fontupper=\small\ttfamily,
  before upper={\raggedright},
}

\title{ObfusQAte: A Proposed Framework to Evaluate LLM Robustness on Obfuscated Factual Question Answering}

\name{Shubhra Ghosh$^{1, *}$, Abhilekh Borah$^{2, *}$, Aditya Kumar Guru$^{2, *}$, Kripabandhu Ghosh$^{3}$}
\address{$^{1}$Indian Institute of Technology Patna, India, 
$^{2}$Manipal University Jaipur, India, \\
$^{3}$
Indian Institute of Science Education and Research Kolkata, India \\
\texttt{shubhra\_2303res116@iitp.ac.in} \\
\texttt{\{abhilekh.229301149, aditya.23fe10cds00348\}@muj.manipal.edu} \\
\texttt{kripaghosh@iiserkol.ac.in}}

\abstract{
The rapid proliferation of \text{Large Language Models (LLMs)} has significantly contributed to the development of equitable AI systems capable of factual question-answering (QA). However, no known study tests the LLMs' robustness when presented with obfuscated versions of questions. To systematically evaluate these limitations, we propose a novel technique, \textbf{\texttt{ObfusQAte}} and leveraging the same, introduce \textbf{ObfusQA}\footnotemark[3], a comprehensive, first of its kind, framework, with \textit{multi-tiered obfuscation levels} designed to examine LLM capabilities across three distinct dimensions: \textbf{(i) Named-Entity Indirection}, \textbf{(ii) Distractor Indirection}, and \textbf{(iii) Contextual Overload}. By capturing these fine-grained distinctions in language, ObfusQA provides a comprehensive benchmark for evaluating LLM robustness and adaptability. Our study observes that LLMs exhibit a tendency to fail or generate hallucinated responses when confronted with these increasingly nuanced variations. To foster research in this direction, we make \href{https://github.com/DeityAG/ObfusQAte/tree/main}{ObfusQAte} publicly available.
\\ \newline 
}

\begin{document}
\maketitleabstract

\renewcommand{\thefootnote}{\fnsymbol{footnote}}
\footnotetext[1]{These authors contributed equally to this work.}
% \footnotetext[2]{\url{https://github.com/DeityAG/ObfusQAte}}
% \footnotetext[3]{\url{https://huggingface.co/datasets/Adignite/ObfusQA}}
% \renewcommand{\thefootnote}{\fnsymbol{footnote}}
% \footnotetext[1]{These authors contributed equally to this work.}

\begin{figure}[h]
    \centering
    \includegraphics[width=\linewidth]{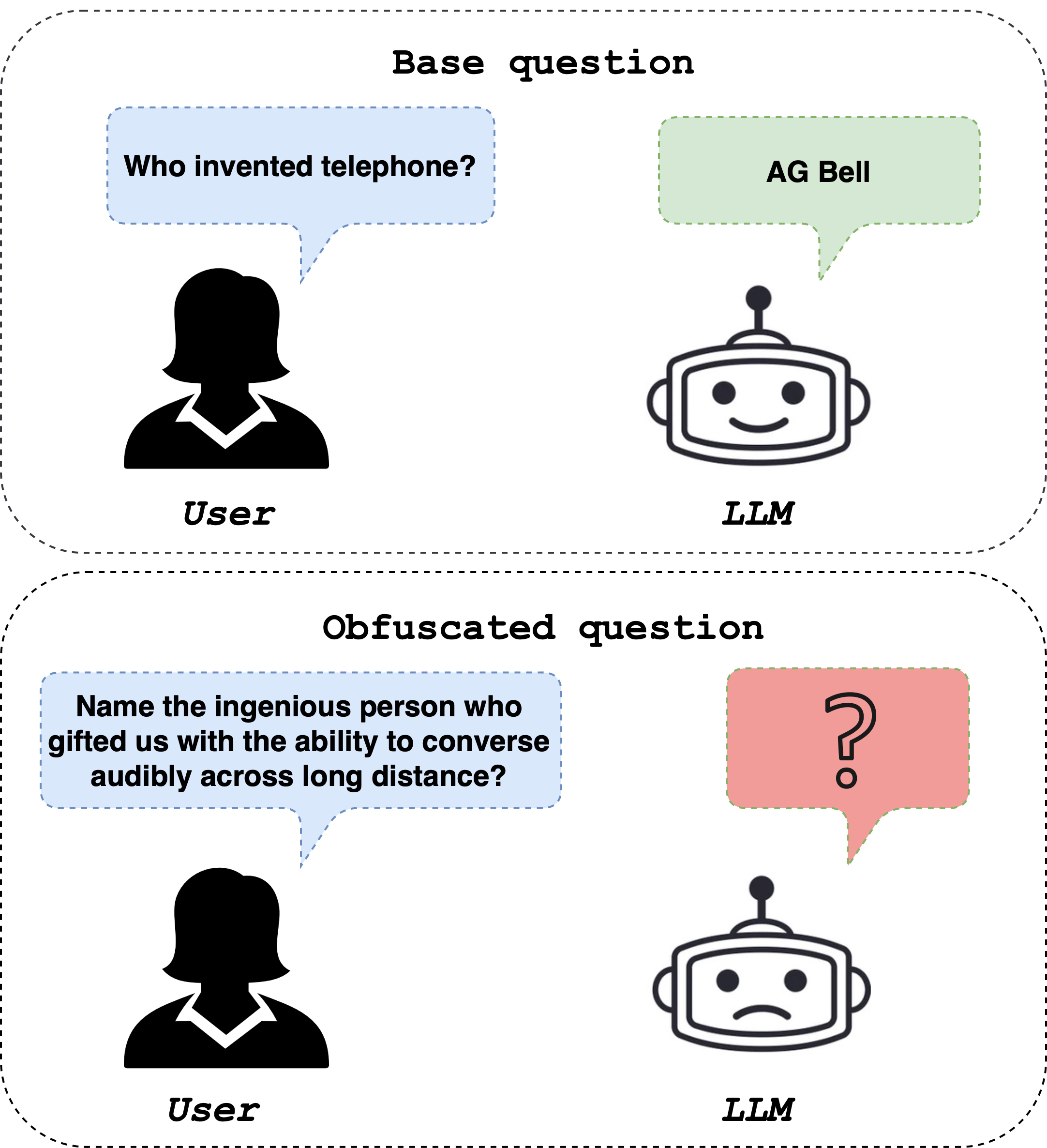}
    \caption{An example of Obfuscated QA between a human and an LLM. Obfuscation in the Question is expected to pose challenges for the LLM even for straightforward questions.}
    \label{fig:obfus-eg}
    \vspace{-4mm}
\end{figure}

\section{Introduction}

In recent times, the Large Language Models (LLMs) like GPT~\cite{achiam2023gpt}, LLaMA~\cite{touvron2023llama}, DeepSeek~\cite{bi2024deepseek} have emerged as game-changers, show-casing unprecedented capabilities of generating coherent responses to a variety of prompts. These models have been applied to numerous tasks, such as report generation, virtual assistants, and summarization, to name a few {\cite{manakul2023selfcheckgpt}.
%Essentially the core large language model is a autoregressive token generation system, which generates next token based on the current context.
Despite their efficacy, these models are plagued by their tendency to generate factually incorrect information with a tone of confidence often termed as {\it hallucination} \cite{azaria2023internal}. The issue of hallucination critically hampers reliability and limits widespread adoption in real-world applications. %Though this phenomenon of hallucination is well known, systematic studies to evaluate hallucination are few like \cite{manakul2023selfcheckgpt, azaria2023internal}.

%Evaluation is of paramount importance to the success of LLMs, particularly for ensuring their safety and reliability in broader applicability areas. 
Based on the study by \citet{chang2024survey}, the current evaluation systems are categorized in many different ways, {\it factuality} being one of them. Factuality in the context of LLMs refers to the extent to which the information or answers provided by the model align with real-world truths and verifiable facts. Factuality in LLMs significantly impacts a variety of tasks and downstream applications, such as QA systems, information extraction, text summarization, dialogue systems, and automated fact-checking, where incorrect or inconsistent information could lead to substantial misunderstandings and misinterpretations. Therefore, evaluating factuality is critical to ensure trust in these models. This includes the ability of these models to maintain consistency with known facts, avoiding generating misleading or false information (known as ``factual hallucination'').  A range of methodologies have been proposed to measure and improve the factuality of language models~\cite{lin2021truthfulqa, wang2023evaluating, pezeshkpour2023measuring, honovich2022true, manakul2023selfcheckgpt}. 

There have been many attempts to improve or test QA capabilities of LLMs \cite{kamalloo2024towards, NEURIPS2023QA}, however, to our knowledge {\it there exists no study that tests the LLMs' ability to perceive semantically obfuscated or obscured variants of an otherwise straightforward question }  (cf. \ref{A.8} for more related works). The analogy is: in an interview, the expert panel tests the knowledge depth of the candidate through intricately nuanced and yet equivalent variants of a potentially straightforward question, for which the candidate might have prepared or even \textit{memorized} in advance. Fig. \ref{fig:obfus-eg} depicts such a scenario, in the case of LLM evaluation.

%While these methods shows an impressive approach to evaluation, they are primarily focused on assessing \textbf{passive recall} {\color{red}KRIPA: cite} of the model and often fail to assess critical cognitive skills such as inferencing and understanding. Inferencing refers to the ability to deduce logically, make logical connections, and retrieval of the correct answer in heavy contextual ambiance obscuring the directness of the question a crucial skill for real-world problem-solving.
%For example in Legal Assistance Systems Legal texts are often complex, and extracting accurate information requires understanding the context. Similary medical professionals rely on accurate, context-aware data to diagnose, recommend treatments, and understand patient histories. {\color{red}KRIPA: cite}
%This gap highlights the need for more robust evaluation frameworks that focus on inference generation, contextual awareness and ambiguity handling. To address this problem, we took inspiration from real life, where prosecutors often ask the same question by adding linguistic nuance to probe the factuality of statements. As fact does not change with linguistic nuance, this is a useful way to probe. In this work, we apply this hypothesis to evaluate factuality of LLMs.
%{\color{red}KRIPA: incomplete sentence}
\begin{figure}
    \centering
    \includegraphics[width=\linewidth]{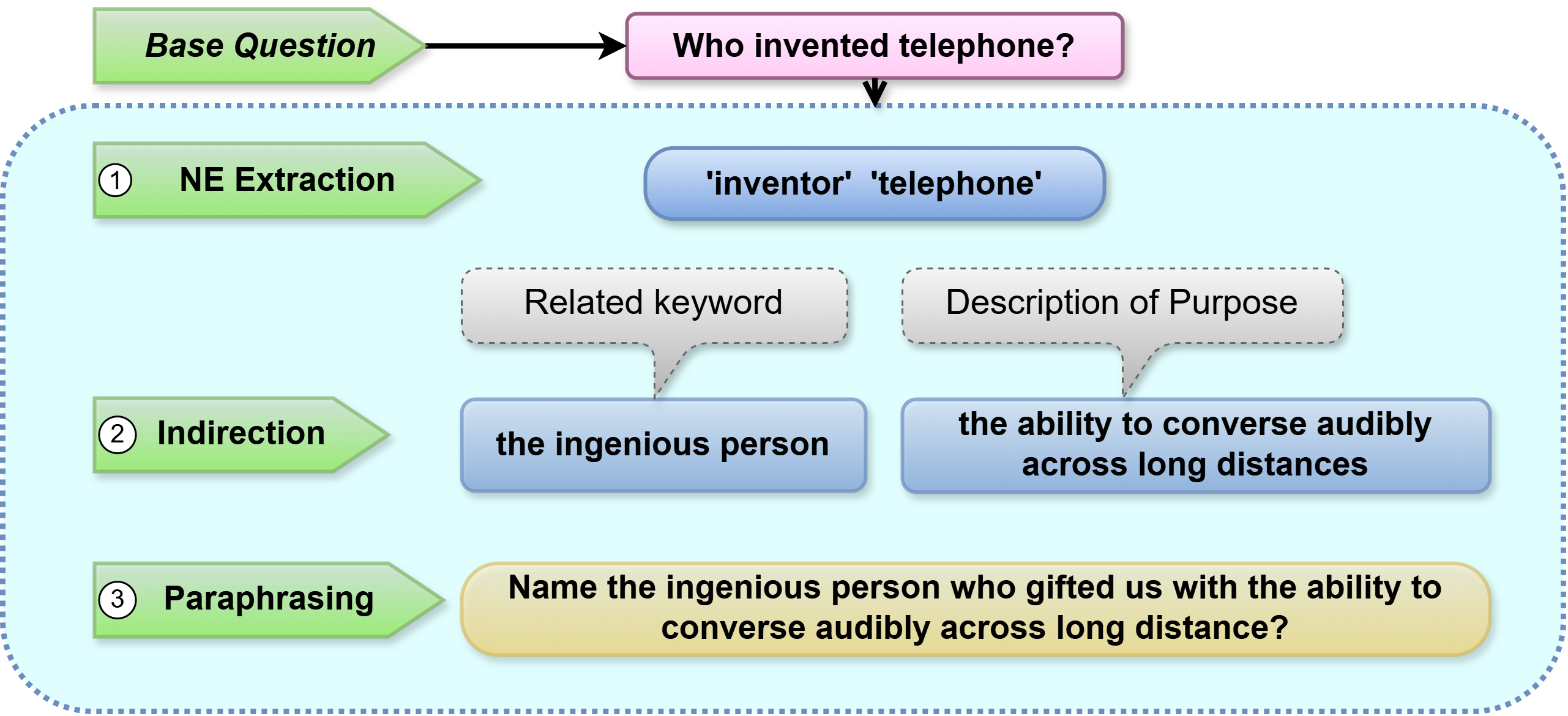}
    \caption{Named-Entity Indirection}
    \label{fig:NEI}
\end{figure}

To address this gap we propose a suite of techniques: \texttt{ObfusQAte} (Sec. \ref{sec:obfusQAte}) leading to obfuscated dataset \textbf{ObfusQA} (Sec. \ref{sec:obfusQA}), a comprehensive, first of its kind, framework, with \textit{multi-tiered obfuscation levels} designed to examine LLM capabilities across three distinct dimensions:  \textbf{(i) Named-Entity Indirection}, \textbf{(ii) Distractor Indirection}, and \textbf{(iii) Contextual Overload}. Finally, we empirically benchmark the efficacy of state-of-the-art LLMs on our proposed setup (Sec. \ref{sec:evaluation}).

% \noindent Our contributions are as follows :
% \begin{itemize}
%     \item Design an Obfuscation algorithm \texttt{ObfusQAte}.
%     \item We create \textbf{ObfusQA} dataset with varying degrees of difficulty.
%     \item We evaluate SOTA LLMs on the \textbf{ObfusQA}.
% \end{itemize}
%We design an Obfuscation algorithm to generate questions characterized by Named entity indirection, plausible distractors, and contextual
%complexities without hindering the actuality.
% \section{Proposed setup: ObfusQAte} \label{sec:obfusQAte}
% To effectively evaluate the capabilities of large language models (LLMs) across a diverse set of challenges, we establish a robust experimental framework built around the \texttt{ObfusQAte} technique. This framework leverages a comprehensive, \textit{multi-tiered obfuscation process} to generate a diverse range of questions that are potentially more complex and challenging than a {\it base question}, while maintaining the semantic essence and the expected answer.

\begin{figure*}[t]
    \centering
    \begin{minipage}{0.48\linewidth}
        \centering
        \includegraphics[width=\linewidth]
        {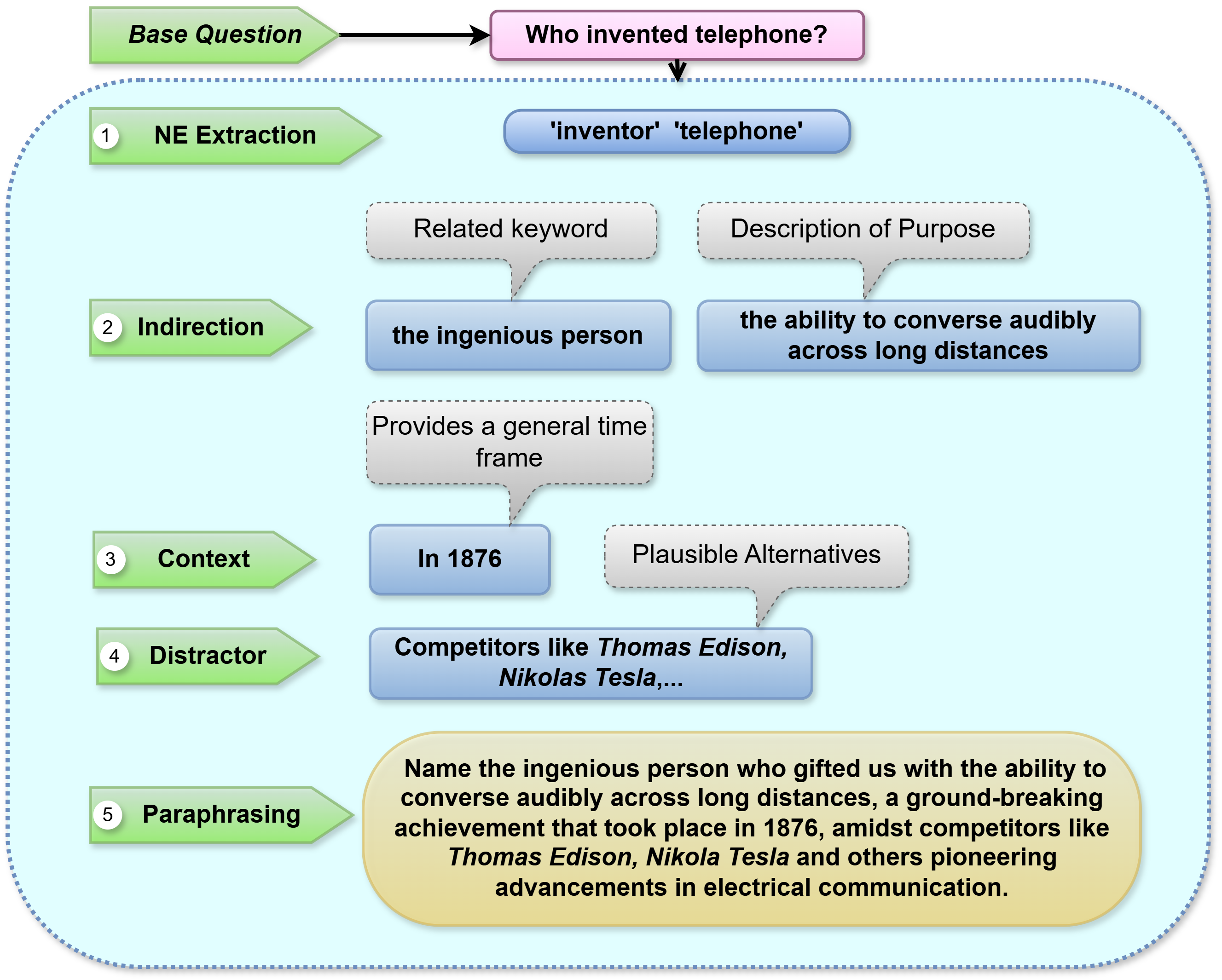}
        \caption{Distractor Indirection}
        \label{fig:DI}
    \end{minipage}
    \hfill
    \begin{minipage}{0.48\linewidth}
        \centering
        \includegraphics[width=0.95\linewidth]
        {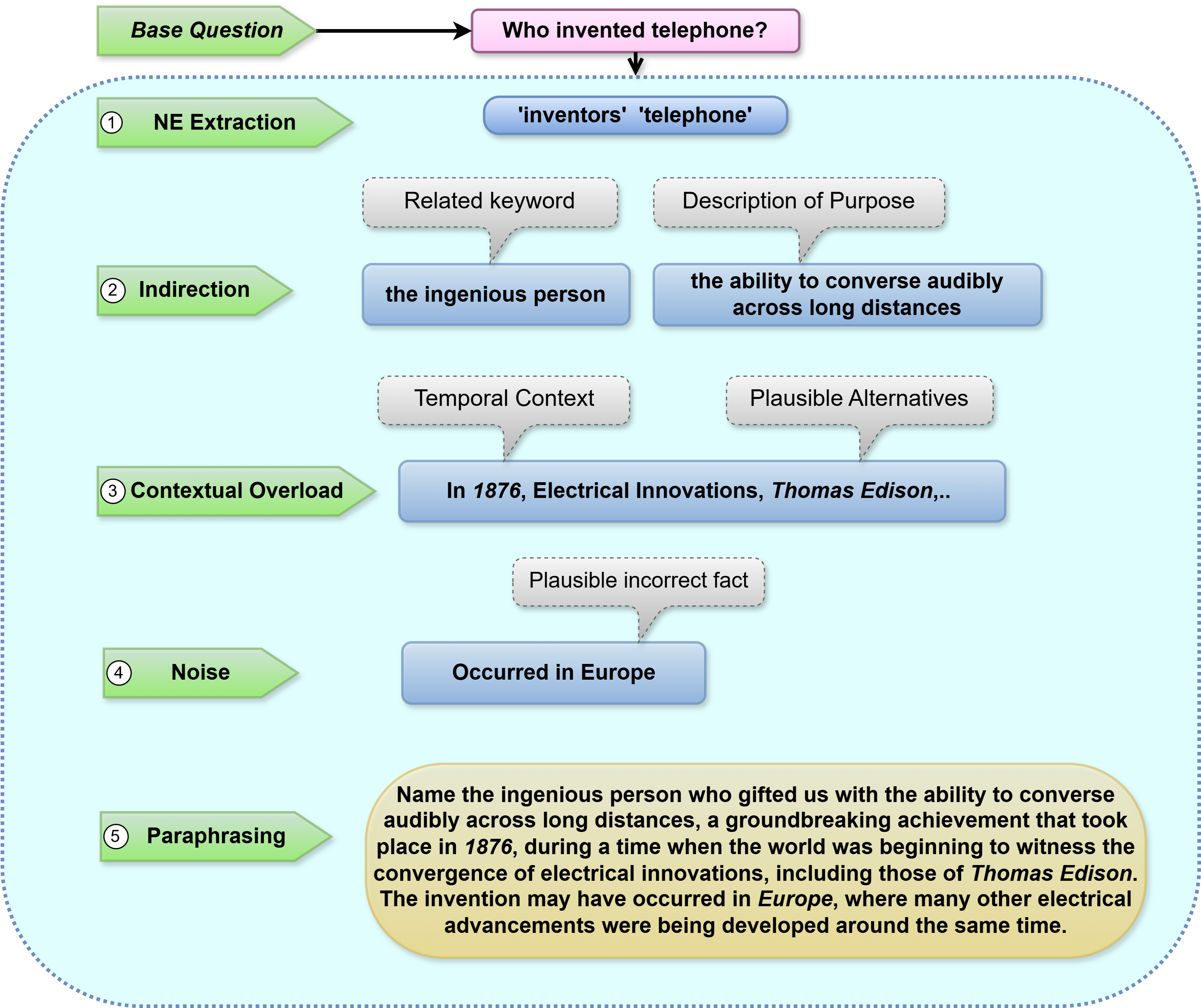}
        \caption{Contextual Overload}
        \label{fig:CO}
    \end{minipage}
    \vspace{-2mm}
\end{figure*}

\subsection{Related Work}
\label{A.8}
Although obfuscation has been discussed in recent LLM studies, very few have systematically assessed its impact on LLM performance.\cite{mohseni2025can} investigates LLMs' potential to generate obfuscated assembly code, developing the MetamorphASM benchmark to evaluate this capability across various models. \cite{swindle2024evaluation} explore how effectively LLMs can detect and analyze obfuscated C++ code, with results highlighting that LLMs struggle particularly with layered and insertion-based obfuscations, revealing a significant gap in their code reasoning abilities. \cite{patsakis2024assessing} investigates LLMs' ability to de-obfuscate malicious PowerShell scripts in real-world malware like Emotet, showing promising results even without specialized training.\cite{pape2025prompt} prompt obfuscation investigates methods for protecting system-level instructions by constructing alternative system prompts that preserve behavioral functionality while rendering the original prompt unintelligible if extracted. It primarily focuses on safeguarding intellectual property against prompt extraction attacks.More recently, \cite{borah2026logiqate} introduced structure-preserving logical obfuscation across first-order logic, blood-relational chains, number series, and direction reasoning, showing that logically equivalent transformations substantially degrade LLM performance despite preserved formal semantics.Jailbreak research such as the WordGame attack demonstrates\cite{zhang-etal-2025-wordgame} how dual obfuscation in both queries and responses can systematically evade safety alignment, effectively bypassing guardrails in leading LLMs and highlighting persistent vulnerabilities in aligned systems.

\section{Proposed setup: ObfusQAte} \label{sec:obfusQAte}
To effectively evaluate the capabilities of LLMs across a diverse set of challenges, we establish a robust experimental framework built around the \texttt{ObfusQAte} technique. This framework leverages a comprehensive, \textit{multi-tiered obfuscation} process to generate a diverse range of questions that are potentially more complex and challenging than a {\it base question} while maintaining the semantic essence and the expected answer.
The generated are stored in the \textbf{ObfusQA \footnote[2]{https://huggingface.co/datasets/Adignite/ObfusQA}} dataset (to be discussed in Section \ref{sec:obfusQA}), which introduces obfuscations along three critical dimensions, as follows:%{\bf Named-Entity Indirection}, {\bf Distractor Indirection} and {\bf Contextual Overload}. Each of these dimensions addresses a key aspect of LLM performance, allowing us to assess how well LLMs handle factual questions transformed with linguistic nuance. Below, we elaborate on the core components of the proposed setup and the motivation behind each aspect.
\vspace{2mm}

\noindent {\bf Named-Entity Indirection (NEI)}  or \textit{Reasoning Through Indirect References}: The motivation behind introducing Named-Entity Indirection is to push LLMs towards deeper, more sophisticated reasoning by forcing them to infer relationships and entities from indirect or abstract cues. Rather than relying on explicit references, this approach challenges the model to connect disparate pieces of information through logical inferences. As shown in Fig.\ref{fig:NEI}, a simple question like \textit{``Who invented the telephone?''} is transformed to \textit{``Name the ingenious person who gifted us with the ability to converse audibly across long distance?''}. In this version, the model after named-entity (NE) extraction, must infer the {\it ``inventor''} by connecting the concept of {\it ``distant audible conversation''} with historical developments in communication technology. The model needs to deduct from the abstract idea of {\it distant audible conversation} to the {\it telephone} specifically. The question might then include additional references to related technologies—like the telegraph, wireless radios, and other communication innovations—which the model must logically connect to arrive at the correct answer. It is important to note that NEI involves reasoning not only about the entities within the question but also those expected in the answer. This process tests the model's ability to make nuanced connections and employs deeper inferencing (e.g., linking  ``telephone'' to ``ability to converse audibly across long distances'') rather than relying on simple memorization of facts.
\vspace{2mm}

\noindent \textbf{Distractor Indirection (DI)} or \textit{ \textit{Actively Steering Toward Wrong Answers}}: Distractor Indirection introduces plausible but incorrect alternatives to steer the model toward false choices deliberately. When a question is framed using both \textbf{indirect references} and \textbf{distractions}, it becomes significantly more obscure by introducing multiple layers of information that include convincing yet incorrect options. This approach tests how well the model can distinguish between similarly plausible answers and how effectively it can sift through distracting, but related, information.For example, in Figure \ref{fig:DI}, the question could be framed as:  \textit{``Name the ingenious person who gifted us with the ability to converse audibly across long distances, a groundbreaking achievement that took place in 1876, amidst competitors like Thomas Edison, Nikola Tesla, and others pioneering advancements in electrical communication.''}
%\textit{How It Works:}
We aim to introduce incorrect but plausible options, forcing the respondent to compare the actual inventor (Alexander Graham Bell) with well-known alternatives -- ``Thomas Edison'' (renowned for electrical innovations) and ``Nikola Tesla'' (associated with pioneering electrical work). The presence of multiple figures in communication technology compels deeper reasoning, making the correct answer less immediately apparent. The correct answer is deliberately de-emphasized, while misleading alternatives seem equally viable, creating a choice dilemma.
\vspace{2mm}

\noindent \textbf{Contextual Overloading (CO)} or \textit{Drowning the Core Question in Noise}: In the contextual overloading frame, we amplify the cognitive load by strategically incorporating \textbf{red herring facts} \footnote{as defined in the \textit{Cambridge Dictionary}, red herring is a  "fact, idea, or subject that takes people's attention away from the central point being considered"} within a heavily overloaded contextual ambiance. In contrast to DI, CO does not steer toward wrong answers but rather buries the correct one under a heavily overloaded contextual environment. This method adds layers of potentially misleading yet related information and adds noises that demand careful reasoning.\\  

For example, Figure \ref{fig:CO} illustrates how a simple base question can be transformed into a heavily contextual one: \textit{``Name the ingenious person who gifted us with the ability to converse audibly across long distances, a groundbreaking achievement that took place in 1876, during a time when the world was beginning to witness the convergence of electrical innovations, including those of Thomas Edison. The invention may have occurred in Europe, where many other electrical advancements were developed around the same time.''} 
Contextual overload achieves the following: \textit{Inject irrelevant but true information} (``the convergence of electrical innovations'', ``electrical advancements in Europe''): forcing the respondent to waste cognitive effort sorting signals from noise.
\textit{Adds excessive but factual complexity} (mentioning Thomas Edison, framing the invention within a global technological shift); making it harder to extract the essential clue.
While all three introduce complexity, they do so in distinct ways: \textbf{NEI} \textit{by requiring the respondent to uncover the intended entity through abstract reasoning and indirect linguistic cues}, \textbf{DI} \textit{by subtly guiding the respondent toward incorrect answers}, and \textbf{CO} \textit{by obscuring the core question with extraneous details}.
 
% \begin{table}[ht]
% \centering
% % \footnotesize
% \adjustbox{max width=\linewidth}{
% \begin{tabular}{l l c c c}
% \toprule
% \textbf{Model} & \textbf{Question Type} & \textbf{Zero-Shot (\%)} & \textbf{Few-Shot (\%)} & \textbf{CoT (\%)} \\
% \midrule
% \multirow{4}{*}{GPT 4o} 
%   & Base   & 67.97 & 73.05 & \textbf{84.38} \\
%   & NEI    & 46.48 & 53.91 & \textbf{55.86} \\
%   & DI     & 25.78 & \textbf{33.59} & 32.42 \\
%   & CO     & 30.08 & 37.89 & \textbf{38.67} \\
% \midrule
% \multirow{4}{*}{Claude 3.5 Sonnet} 
%   & Base   & \textbf{78.91} & 74.61 & 75.00 \\
%   & NEI    & 36.72 & 41.41 & \textbf{54.30} \\
%   & DI     & 26.17 & 27.73 & \textbf{38.28} \\
%   & CO     & 35.16 & 32.81 & \textbf{39.45} \\
% \midrule
% \multirow{4}{*}{DeepSeek R1} 
%   & Base   & 71.25 & 76.80 & \textbf{82.15} \\
%   & NEI    & 42.30 & 48.67 & \textbf{58.92} \\
%   & DI     & 28.45 & 31.20 & \textbf{40.78} \\
%   & CO     & 33.90 & 39.15 & \textbf{42.33} \\
% \midrule
% \multirow{4}{*}{o3-mini} 
%   & Base   & 69.80 & \textbf{79.60} & 72.45 \\
%   & NEI    & 40.85 & \textbf{52.75} & 45.20 \\
%   & DI     & 27.60 & 30.85 & \textbf{36.90} \\
%   & CO     & 32.40 & \textbf{40.25} & 36.70 \\
% \bottomrule
% \end{tabular}
% }
% \vspace{0.5em}
% \caption{EM (\%) accuracy of GPT-4o and Claude 3.5 Sonnet across obfuscation types under zero-shot, few-shot, and CoT prompting. GPT-4o benefits from CoT, while Claude 3.5 often leads in zero-shot. Reasoning models (DeepSeek R1, o3-mini) show drops on obfuscated inputs. \textbf{Bold} entries indicate row-wise best performance.}
% \label{tab:evaluation_results}
% \end{table}

\begin{figure}[t]
    \centering
    \includegraphics[width=0.9\linewidth]{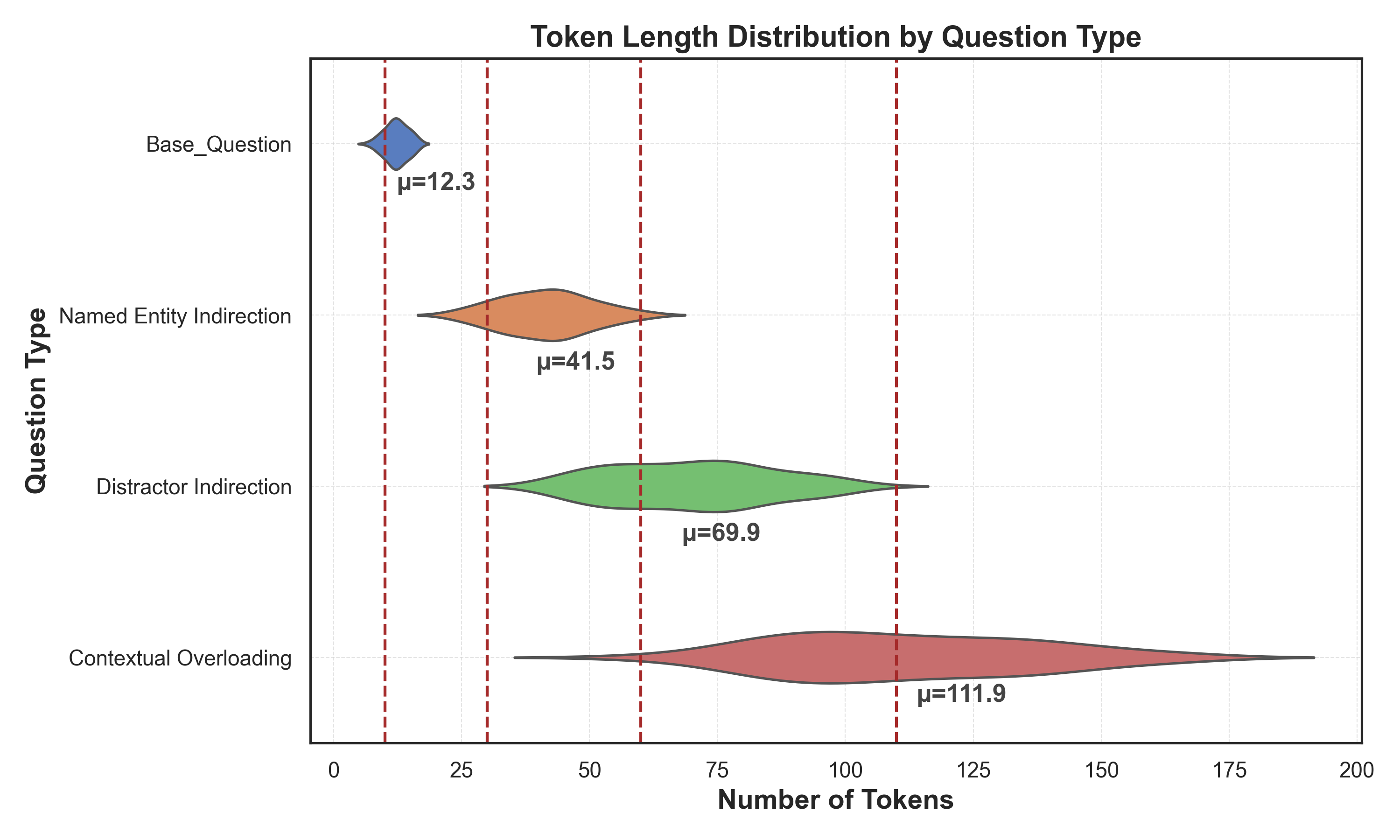}
\caption{Token length distributions across ObfusQA types. Complexity increases with obfuscation; red dashed lines denote mean (\(\mu\)) length.}
    \label{fig:token-lengths}
    \vspace{-5mm}
\end{figure}
% \vspace{-1mm}

\subsection{Dataset Creation: ObfusQA}\label{sec:obfusQA}
We source our base questions primarily from the TriviaQA dataset~\cite{joshi-etal-2017-triviaqa}, along with a small subset drawn from a government examination preparation website, GKToday\footnote{\url{https://www.gktoday.in/}}. Following this, we utilize Gemini~2.0 Flash~\cite{gemini2.0, team2024gemini} to generate obfuscations from the base questions using our designed algorithm through well-structured prompts. All generations were conducted at a temperature of 0.75. After obfuscating these base questions, we obtain a total of 1024 questions that include both the base questions and their three obfuscations, verified and corrected through a \textit{human-in-the-loop} process by annotators.

A team of seven undergraduate annotators from an Indian university’s AI research lab, with prior coursework and experience in NLP, manually reviewed each generated question. They interpreted, analyzed, and refined the model outputs to correct hallucinations and maintain semantic faithfulness to the original query. Human annotation safeguarded dataset integrity by rigorously overseeing each transformation to ensure both factual consistency and semantic alignment. Annotators focused on two key principles: \textbf{(i) Ground Truth Preservation}, ensuring that every obfuscated question continued to map unambiguously to the same factual answer, and \textbf{(ii) Cognitive Load without Ambiguity}, ensuring that added linguistic or contextual complexity increased reasoning difficulty without introducing factual drift. For instance, the base question \textit{``What is the capital of Australia?''} was transformed into its obfuscated variants: (a) \text{Named-Entity Indirection (NEI)}: \emph{``Which urban center, situated within the Commonwealth realm that witnessed the dawning of the new millennium with a quadrennial celebration of athletic prowess, serves as the locus of governance for a continent-spanning island nation?''}; (b) \text{Distraction Indirection (DI)}: \emph{``Amidst ongoing debates about regional development, is the principal federal city—let's call it X—more populous than the metropolis that annually celebrates equestrian prowess, or does it rival the harbour city known for its architectural marvel?''}; and (c) \text{Contextual Overload (CO)}: \emph{``Amidst the echoes of the Great Emu War and the ongoing debate over the Pavlova's origins, can you identify the city, nestled within the Australian Capital Territory, that serves as the seat of the Governor-General and houses the Old Parliament House near Lake Burley Griffin?''}. All variants retained the same \text{ground truth answer:} \emph{``Canberra''}. This process exemplifies how annotators validated semantic equivalence and factual correctness while allowing cognitive difficulty to scale with obfuscation level. Disagreements among annotators were resolved through adjudication (e.g., when two annotators provided differing interpretations of an obfuscated question’s intent or correctness, a third reviewer compared both annotations against the original base question and final ground truth to determine the most semantically faithful version) to ensure consistency and high-quality supervision across the dataset.

To assess annotation reliability, we computed \textit{Cohen’s Kappa} ($\kappa$) between two independent annotators across all variants:
\[
\kappa = \frac{p_o - p_e}{1 - p_e}
\]
where $p_o$ denotes observed agreement and $p_e$ represents expected agreement by chance. The resulting $\kappa = 0.862$ indicates \textit{strong agreement} (86.2\%) under standard interpretation thresholds. 
The detailed prompt designs, annotation workflow and adjudication protocol are described comprehensively in Appendix ~\ref{sec:obfusprompts} and ~\ref{A.6}.

An analysis of the average token lengths for each obfuscation type in ObfusQAte further confirms the designed progression in verbosity and reasoning load: base questions average 11.6 tokens, NEI variants 41.9, DI 62.3, and CO variants 116.1 tokens (cf.\ Figure~\ref{fig:token-lengths}).

\section{Evaluation Setup}\label{sec:evaluation}
\vspace{-1mm}

We benchmarked seven SoTA LLMs on ObfusQA: GPT-4o \cite{hurst2024gpt}, GPT-4o mini \cite{openai24}, LLaMA 3.3 70b \cite{dubey2024llama}, Gemini 2.0 Flash, Claude 3.5 Sonnet \cite{claudesonnet}; we include two strong \textit{reasoning} models: DeepSeek R1 \cite{bi2024deepseek}, and GPT o3-mini \cite{openaio3}. We evaluated performance using \textbf{Exact Match (EM) accuracy}, which measures the percentage of normalized outputs exactly matching a normalized ground truth answer.

\vspace{-2mm}

\subsection{Evaluation Metric}
\label{A.2}
We evaluate our system using an \textbf{Exact Match (EM) Accuracy} metric, which quantifies the percentage of samples for which the normalized model answer exactly matches one of the normalized ground truth answers. For example, given the ground truth ``Pope'', a generated answer like ``Pope, and his relative'' would be considered incorrect due to the exact match requirement. Mathematically, let \(N\) be the total number of samples. For each sample \(i\), let \(\mathcal{Y}^{(i)}\) be the set of normalized ground truth answers and \(\hat{y}^{(i)}\) the normalized answer produced by the model. We define an indicator function \(I^{(i)}\) such that \(I^{(i)}=1\) if \(\hat{y}^{(i)} \in \mathcal{Y}^{(i)}\) and \(I^{(i)}=0\) otherwise. The overall EM accuracy is then computed as:

\begin{equation}
\text{EM Accuracy} = \frac{1}{N} \sum_{i=1}^{N} I^{(i)} \times 100\%.
\end{equation}

For each of the obfuscation variant, the same methodological formulation is applied. The normalization process standardizes case, removes punctuation, and ensures whitespace uniformity, making comparisons resilient to peripheral textual variations.
\begin{table}[ht]
\centering
\adjustbox{max width=\linewidth}{
\begin{tabular}{l l c c c}
\toprule
\textbf{Model} & \textbf{Question Type} & \textbf{Zero-Shot (\%)} & \textbf{Few-Shot (\%)} & \textbf{CoT (\%)} \\
\midrule
\multirow{4}{*}{DeepSeek R1} 
  & Base   & 71.25 & 76.80 & \textbf{82.15} \\
  & NEI    & 42.30 & 48.67 & \textbf{58.92} \\
  & DI     & 28.45 & 31.20 & \textbf{40.78} \\
  & CO     & 33.90 & 39.15 & \textbf{42.33} \\
\midrule
\multirow{4}{*}{o3-mini} 
  & Base   & 69.80 & \textbf{79.60} & 72.45 \\
  & NEI    & 40.85 & \textbf{52.75} & 45.20 \\
  & DI     & 27.60 & 30.85 & \textbf{36.90} \\
  & CO     & 32.40 & \textbf{40.25} & 36.70 \\
\bottomrule
\end{tabular}
}
\vspace{0.5em}
\caption{EM (\%) accuracy of DeepSeek R1 and o3-mini across obfuscation types (100 samples). Both models benefit from CoT prompting; DeepSeek R1 showing stronger CoT gains on obfuscated inputs.}
\label{tab:deepseek_o3_comparison}
\vspace{-4mm}
\end{table}
\subsection{Results and Analysis}
We evaluate the LLMs across \textit{zero-shot}, \textit{few-shot}, and \textit{chain-of-thought (CoT)} prompting strategies to measure their robustness under varying reasoning settings (cf. Table \ref{tab:combined_results} and \ref{tab:deepseek_o3_comparison}). While models perform well on base questions, their accuracy on its obfuscated counterparts, particularly for \text{Distractor Indirection (DI)} and \text{Contextual Overload (CO)} variants, which demand higher cognitive reasoning and disambiguation. GPT-4o shows an average performance degradation of \text{56\%} from its base performance, Claude 3.5~Sonnet drops by \text{49\%}, LLaMA~3.3~70B by \text{44\%}, GPT-4o~mini by \text{57\%}, and Gemini~Flash~2.0 by \text{55\%}. This performance decline highlights the difficulty current models face when interpreting semantic cues (e.g., indirect entity references such as “the ingenious person who gifted us with the ability to converse audibly across long distances” instead of “the inventor of the telephone”) embedded within increasingly convoluted linguistic structures (lengthy contextual descriptions or distractor clauses that obscure the main intent of the question), which collectively disrupt direct pattern retrieval from their memorized associations and training data representations.

We also evaluated \text{Gemini~2.0~Flash} to assess its ability to answer its own, generated obfuscated questions. Despite being the source model for query generation, Gemini~2.0~Flash failed to correctly answer most transformed queries, revealing limitations in model \textit{“self-awareness”} and resilience to its own obfuscations. Due to budget constraints, 100 samples were used to evaluate reasoning-oriented models (\text{DeepSeek~R1} and \text{GPT~o3-mini}), which showed reduced performance on obfuscated inputs. DeepSeek~R1, experienced a degradation of \text{50\%} from Base (82.2\%) to DI (40.8\%), while o3-mini dropped by approximately \text{49\%}, confirming a consistent vulnerability across reasoning-focused models. Across all models, \text{Chain-of-Thought (CoT)} prompting consistently improves accuracy by an average of \text{8-12\%} compared to few-shot settings, demonstrating that explicit reasoning steps help models better parse layered or indirect phrasing. In contrast, \text{Few-Shot} prompting provides only marginal improvements of around \text{2-4\%} and occasionally results in negligible or negative gains (Claude~3.5~Sonnet’s base accuracy decreases slightly from 78.9\% to 74.6\%), suggesting that exposure to exemplars alone does not sufficiently enhance generalization under obfuscation. \text{Zero-Shot} performance remains the weakest, averaging about \text{19\% lower} than CoT and roughly \text{11\% lower} than few-shot across all models. Overall, while most models demonstrate robust factual retention on base questions, their ability to generalize to obfuscated variants decreases drastically as linguistic complexity and distractor density increase.

\begin{table*}[ht]
\centering
\adjustbox{max width=\linewidth}{
\begin{tabular}{l l c c c}
\toprule
\textbf{Model} & \textbf{Question Type} & \textbf{Zero-Shot (\%)} & \textbf{Few-Shot (\%)} & \textbf{CoT (\%)} \\
\midrule
\multirow{4}{*}{GPT-4o} 
  & Base   & 67.97 & 73.05 & \textbf{84.38} \\
  & Named-Entity Indirection    & 46.48 & 53.91 & \textbf{55.86} \\
  & Distractor Indirection     & 25.78 & \textbf{33.59} & 32.42 \\
  & Contextual Overload     & 30.08 & 37.89 & \textbf{38.67} \\
\midrule
\multirow{4}{*}{Claude 3.5 Sonnet} 
  & Base   & \textbf{78.91} & 74.61 & 75.00 \\
  & Named-Entity Indirection    & 36.72 & 41.41 & \textbf{54.30} \\
  & Distractor Indirection     & 26.17 & 27.73 & \textbf{38.28} \\
  & Contextual Overload     & 35.16 & 32.81 & \textbf{39.45} \\
\midrule
\multirow{4}{*}{LLaMA 3.3 70b} 
  & Base   & 75.69 & \textbf{77.34} & 74.61 \\
  & Named-Entity Indirection  & \textbf{43.14} & 40.23 & 41.41 \\
  & Distractor Indirection  & 29.80 & \textbf{30.08} & \textbf{30.08} \\
  & Contextual Overload  & 32.55 & 32.81 & \textbf{35.55} \\
\midrule
\multirow{4}{*}{GPT-4o mini} 
  & Base   & 57.81 & 57.42 & \textbf{61.72} \\
  & Named-Entity Indirection  & 31.64 & 32.42 & \textbf{36.72} \\
  & Distractor Indirection  & 23.05 & 24.22 & \textbf{26.17} \\
  & Contextual Overload  & 23.44 & 26.95 & \textbf{30.08} \\
\midrule
\multirow{4}{*}{Gemini Flash 2.0} 
  & Base   & 72.27 & 76.95 & \textbf{78.91} \\
  & Named-Entity Indirection  & 44.92 & 48.44 & \textbf{50.78} \\
  & Distractor Indirection  & 32.03 & \textbf{36.72} & 33.59 \\
  & Contextual Overload  & 36.72 & \textbf{37.50} & 35.55 \\
\bottomrule
\end{tabular}
}
\vspace{0.5em}
\caption{Evaluation of GPT-4o, Claude 3.5 Sonnet, LLaMA 3.3 70B, GPT-4o mini, and Gemini Flash 2.0 across obfuscation types under Zero-Shot, Few-Shot, and Chain-of-Thought (CoT) prompting. EM (\%) accuracy values are shown, with \textbf{bold-faced} entries indicating the best performance per row. GPT-4o shows strong CoT reasoning gains, Claude 3.5 excels in Zero-Shot, and Gemini Flash 2.0 maintains consistency across settings.}
\label{tab:combined_results}
\vspace{-4mm}
\end{table*}

\begin{table*}[ht]
\centering
\small
\begin{tabular}{@{}p{1.5cm} p{9cm} p{4cm}@{}}
\toprule
\textbf{Question Type} & \textbf{Token Analysis (\(P(\text{IK})\))} & \textbf{Model Answer (Ground truth: Duck)}\\
\midrule
Base &
\highlighttoken{\texttt{<|begin\_of\_text|>}}{0.159302}\,
\highlighttoken{What}{0.351807}\,
\highlighttoken{sort}{0.995117}\,
\highlighttoken{of}{0.017242}\,
\highlighttoken{creature}{0.478516}\,
\highlighttoken{is}{0.338867}\,
\highlighttoken{a}{0.010796}\,
\highlighttoken{'}{0.020844}\,
\highlighttoken{M}{0.057373}\,
\highlighttoken{and}{0.649536}\,
\highlighttoken{arin}{0.805566}\,
\highlighttoken{'?}{0.077515}
& Duck
\\[1ex]

NEI &
\highlighttoken{\texttt{<|begin\_of\_text|>}}{0.159424}\,
\highlighttoken{What}{0.351807}\,
\highlighttoken{type}{0.985840}\,
\highlighttoken{of}{0.022232}\,
\highlighttoken{being}{0.261475}\,
\highlighttoken{is}{0.238403}\,
\highlighttoken{referred}{0.941895}\,
\highlighttoken{to}{0.465088}\,
\highlighttoken{by}{0.682129}\,
\highlighttoken{a}{0.043427}\,
\highlighttoken{term}{0.199341}\,
\highlighttoken{reminiscent}{0.977051}\,
\highlighttoken{of}{0.205933}\,
\highlighttoken{both}{0.115906}\,
\highlighttoken{a}{0.067566}\,
\highlighttoken{high}{0.144531}\,
\highlighttoken{-ranking}{0.099121}\,
\highlighttoken{bureauc}{0.995605}\,
\highlighttoken{rat}{0.739258}\,
\highlighttoken{in}{0.404541}\,
\highlighttoken{a}{0.173462}\,
\highlighttoken{historical}{0.139648}\,
\highlighttoken{Eastern}{0.309326}\,
\highlighttoken{empire}{0.808105}\,
\highlighttoken{and}{0.567871}\,
\highlighttoken{a}{0.028595}\,
\highlighttoken{brightly}{0.556152}\,
\highlighttoken{waterbird}{0.571387}\,
\highlighttoken{?}{0.138916}
& A Mandarin duck
\\[1ex]

DI &
\highlighttoken{\texttt{<|begin\_of\_text|>}}{0.159424}\,
\highlighttoken{Known}{0.142578}\,
\highlighttoken{for}{0.063477}\,
\highlighttoken{its}{0.098145}\,
\highlighttoken{vibrant}{0.462891}\,
\highlighttoken{plumage}{0.982422}\,
\highlighttoken{,}{0.178223}\,
\highlighttoken{what}{0.273438}\,
\highlighttoken{avian}{0.678223}\,
\highlighttoken{species}{0.798828}\,
\highlighttoken{,}{0.152344}\,
\highlighttoken{often}{0.119141}\,
\highlighttoken{associated}{0.238281}\,
\highlighttoken{with}{0.105469}\,
\highlighttoken{Asian}{0.583984}\,
\highlighttoken{waterways}{0.715820}\,
\highlighttoken{,}{0.138672}\,
\highlighttoken{shares}{0.167969}\,
\highlighttoken{its}{0.093750}\,
\highlighttoken{name}{0.125000}\,
\highlighttoken{with}{0.084961}\,
\highlighttoken{a}{0.051758}\,
\highlighttoken{high-ranking}{0.379883}\,
\highlighttoken{official}{0.492188}\,
\highlighttoken{in}{0.097656}\,
\highlighttoken{imperial}{0.402344}\,
\highlighttoken{China}{0.654297}\,
\highlighttoken{,}{0.193359}\,
\highlighttoken{and}{0.084473}\,
\highlighttoken{is}{0.071289}\,
\highlighttoken{sometimes}{0.209961}\,
\highlighttoken{confused}{0.295898}\,
\highlighttoken{with}{0.102539}\,
\highlighttoken{the}{0.074219}\,
\highlighttoken{closely}{0.129395}\,
\highlighttoken{related}{0.225098}\,
\highlighttoken{"Carolina"}{0.752930}\,
\highlighttoken{variety}{0.449219}\,
\highlighttoken{?}{0.113281}.
& Wigeon
\\[1ex]

CO &
\highlighttoken{\texttt{<|begin\_of\_text|>}}{0.159424}\,
\highlighttoken{In}{0.168701}\,
\highlighttoken{the}{0.050323}\,
\highlighttoken{context}{0.989746}\,
\highlighttoken{of}{0.300293}\,
\highlighttoken{19}{0.831543}\,
\highlighttoken{th}{0.502930}\,
\highlighttoken{-century}{0.068909}\,
\highlighttoken{colonial}{0.469238}\,
\highlighttoken{discourse}{0.729492}\,
\highlighttoken{,}{0.208862}\,
\highlighttoken{what}{0.230713}\,
\highlighttoken{type}{0.983398}\,
\highlighttoken{of}{0.057709}\,
\highlighttoken{waterfowl}{0.390137}\,
\highlighttoken{might}{0.183105}\,
\highlighttoken{one}{0.130981}\,
\highlighttoken{associate}{0.947266}\,
\highlighttoken{with}{0.480469}\,
\highlighttoken{the}{0.079224}\,
\highlighttoken{mon}{0.792969}\,
\highlighttoken{iker}{0.422119}\,
\highlighttoken{ev}{0.502441}\,
\highlighttoken{oking}{0.280518}\,
\highlighttoken{both}{0.185669}\,
\highlighttoken{a}{0.058411}\,
\highlighttoken{specific}{0.047455}\,
\highlighttoken{citrus}{0.753906}\,
\highlighttoken{fruit}{0.812988}\,
\highlighttoken{and}{0.483398}\,
\highlighttoken{a}{0.059753}\,
\highlighttoken{high}{0.105469}\,
\highlighttoken{-ranking}{0.070801}\,
\highlighttoken{official}{0.230225}\,
\highlighttoken{,}{0.129639}\,
\highlighttoken{particularly}{0.228638}\,
\highlighttoken{when}{0.224609}\,
\highlighttoken{referencing}{0.280518}\,
\highlighttoken{descriptions}{0.827637}\,
\highlighttoken{from}{0.282471}\,
\highlighttoken{individuals}{0.193726}\,
\highlighttoken{travers}{0.999023}\,
\highlighttoken{ing}{0.400146}\,
\highlighttoken{Southeast}{0.854004}\,
\highlighttoken{Asia}{0.262939}\,
\highlighttoken{?}{0.088440}
& Orange Rajah
\\[1ex]

\bottomrule
\end{tabular}

% Add some vertical space before the scale
\vspace{1em}

% Put the horizontal scale here, so it appears "in the same image"
\probsclhoriz

\caption{Representation of token-wise \(P(\text{IK})\) scores for various obfuscated query variants used to probe the LLaMA 3.1 8b model. The Base query, with its straightforward phrasing, concentrates activations on key tokens, leading to the correct answer, “Duck”. In the NEI variant, despite added descriptive elements, the essential cues remain sufficiently prominent to yield an answer close to the ground truth ("A Mandarin duck"). However, the DI and CO queries introduce further obfuscation through extra contextual and indirect references, which shift the model’s focus and alter the token activation pattern, ultimately resulting in incorrect responses ("Wigeon" and "Orange Rajah").}
\label{tab:Probing-table}
\end{table*}

\section{Intrinsic Analysis}
\label{A.4}

\vspace{2mm}
\noindent
To better understand how obfuscation affects internal reasoning and representation, we conducted three targeted analyses using \text{LLaMA~3.1~8B}~\cite{dubey2024llama} and \text{Mistral~7B~v0.1}~\cite{jiang2023mistral7b}, constrained by GPU availability. Although both models underperformed on ObfusQA overall, they provide insight into how internal mechanisms: token confidence, memorization, and representational compression, shift with increasing cognitive load. We examined: \textbf{(i) Intrinsic Confidence}, reflected in the model’s self-assessed probability of correctness; \textbf{(ii) Memorization}, via membership inference to assess whether obfuscated variants appeared in pre-training; and \textbf{(iii) Layer-wise Norm Drop Analysis}, revealing when feature compression occurs during forward propagation.

\subsection{Understanding Intrinsic Confidence}
\label{sec:token}
To quantify internal certainty, we evaluate the \(P(\text{IK})\) scores, defined as the probability a model assigns to the phrase “I know” (i.e., its self-estimated likelihood of answering correctly when sampled at unit temperature)~\cite{kadavath2022language} (cf. Table \ref{tab:Probing-table}) . We perform token-level \(P(\text{IK})\) analysis across all obfuscation types. Average \(P(\text{IK})\) drops by roughly \text{28-32\%} for \text{Named-Entity Indirection (NEI)}, \text{42-46\%} for \text{Distractor Indirection (DI)}, and up to \text{51\%} for \text{Contextual Overload (CO)}. This suggests that as linguistic complexity increases (e.g., through layered modifiers, distractors, or redundant clauses), models become less certain about their own correctness, an effect that correlates with their observed EM accuracy declines (cf.\ Table~\ref{tab:combined_results}).
\begin{figure*}[htbp]
    \centering
    \begin{minipage}[t]{0.24\textwidth}
        \centering
        \includegraphics[width=\textwidth, height=4cm]{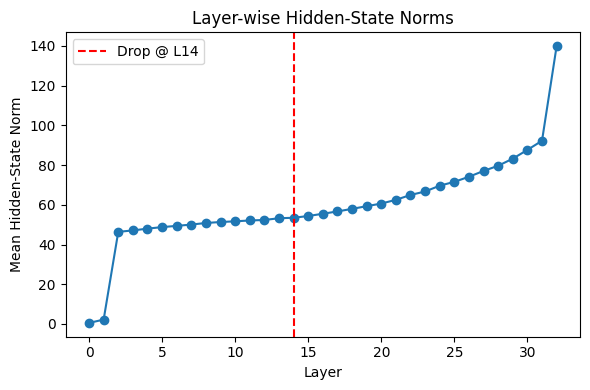}
        
        \vspace{0.3em}
        \textbf{(a)} Base Question
    \end{minipage}
    \hfill
    \begin{minipage}[t]{0.24\textwidth}
        \centering
        \includegraphics[width=\textwidth, height=4cm]{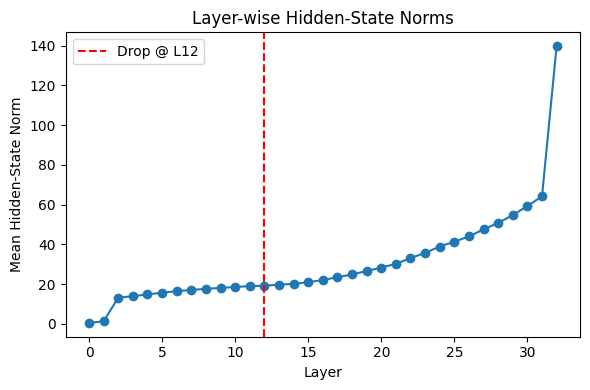}
        
        \vspace{0.3em}
        \textbf{(b)} NEI
    \end{minipage}
    \hfill
    \begin{minipage}[t]{0.24\textwidth}
        \centering
        \includegraphics[width=\textwidth, height=4cm]{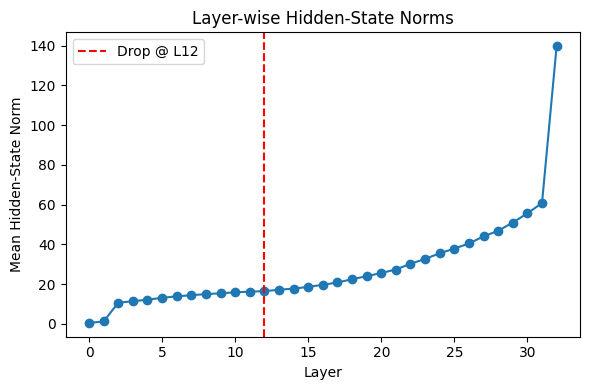}
        
        \vspace{0.3em}
        \textbf{(c)} DI
    \end{minipage}
    \hfill
    \begin{minipage}[t]{0.24\textwidth}
        \centering
        \includegraphics[width=\textwidth, height=4cm]{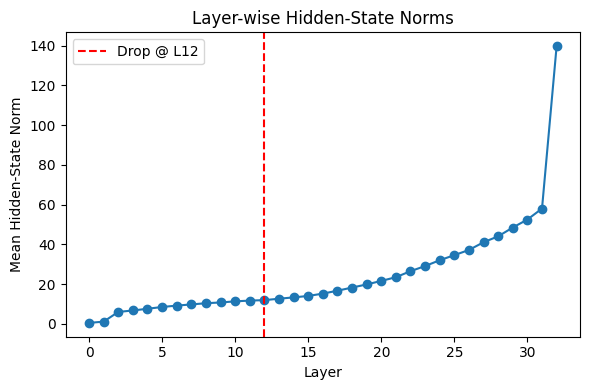}
        
        \vspace{0.3em}
        \textbf{(d)} CO
    \end{minipage}
    
    \caption{Comparison of model responses to four question variants about the largest planet in the solar system. 
    \textbf{(a)} \textit{Base Question}: "What is the largest planet in our solar system?"
    \textbf{(b)} \textit{NEI (Named Entity Indirection)}: "What celestial body, renowned as the most expansive in the star system we call home, would its diameter compare to when measured against the circumference of the third rock from the sun, assuming its equatorial dimension?"
    \textbf{(c)} \textit{DI (Distractor Indirection)}: "Given its gaseous composition and significant influence on neighboring celestial bodies, which orbiter, often referred to as the "king" of our local star's family, surpasses all others in diameter, and is it larger than, say, Neptune or Uranus?"
    \textbf{(d)} \textit{CO (Contextual Overload)}: "Amidst the celestial bodies influenced by the radiant star at the heart of our local cosmic neighborhood, and considering the understanding of astronomical entities held by ancient Greco-Roman civilizations, which gas giant, whose namesake ruled over the gods, surpasses all others in volumetric magnitude, dwarfing even the terrestrial sphere that cradles the city known for the Colosseum, and also has a storm that is larger than three times the area of Earth? Also, is it true that the planet is mostly made of solid rock?" 
    \textbf{Ground truth:} Jupiter.}
    \label{fig:layernorm}
\end{figure*}
\subsection{Memorization}
\label{sec:member}
To test whether our obfuscated questions appeared during model pre-training, we apply the \text{Membership Inference Attack (MIA)} using the \text{Min-K\%++} method~\cite{zhang2025mink}, where \(K\) denotes the percentage of lowest-probability token sequences contributing to the score (cf. Fig \ref{fig:mink++mistral} and \ref{fig:mink++}). Higher Area Under ROC (AUROC) indicates stronger evidence that the queried data overlaps with training samples. 

\paragraph{LLaMA~3.1~8B.} Base questions achieved AUROC between 47-55\%, while DI and CO variants decreased by nearly \text{20\%} (to 38-44\%), showing that obfuscation substantially weakens the statistical alignment between queries and pre-training data. NEI questions maintained moderate detectability (43-45\%), implying that light paraphrasing preserves partial overlap.

\paragraph{Mistral~7B~v0.1.} AUROC for base questions ranged from 53\% (at $K{=}0$) to 50\% (at $K{=}1.0$), while DI and CO dropped by \text{15-18\%}. NEI increased slightly from 42\% to 45\%, consistent with less aggressive linguistic perturbation. Across both models, the general trend indicates that higher obfuscation (especially CO) disrupts token-level recall and reduces pre-training membership detection, confirming that these variants are effectively novel to the models’ learned space.

\begin{figure}[ht]
    \centering
    \includegraphics[width=0.48\textwidth, height=6cm]{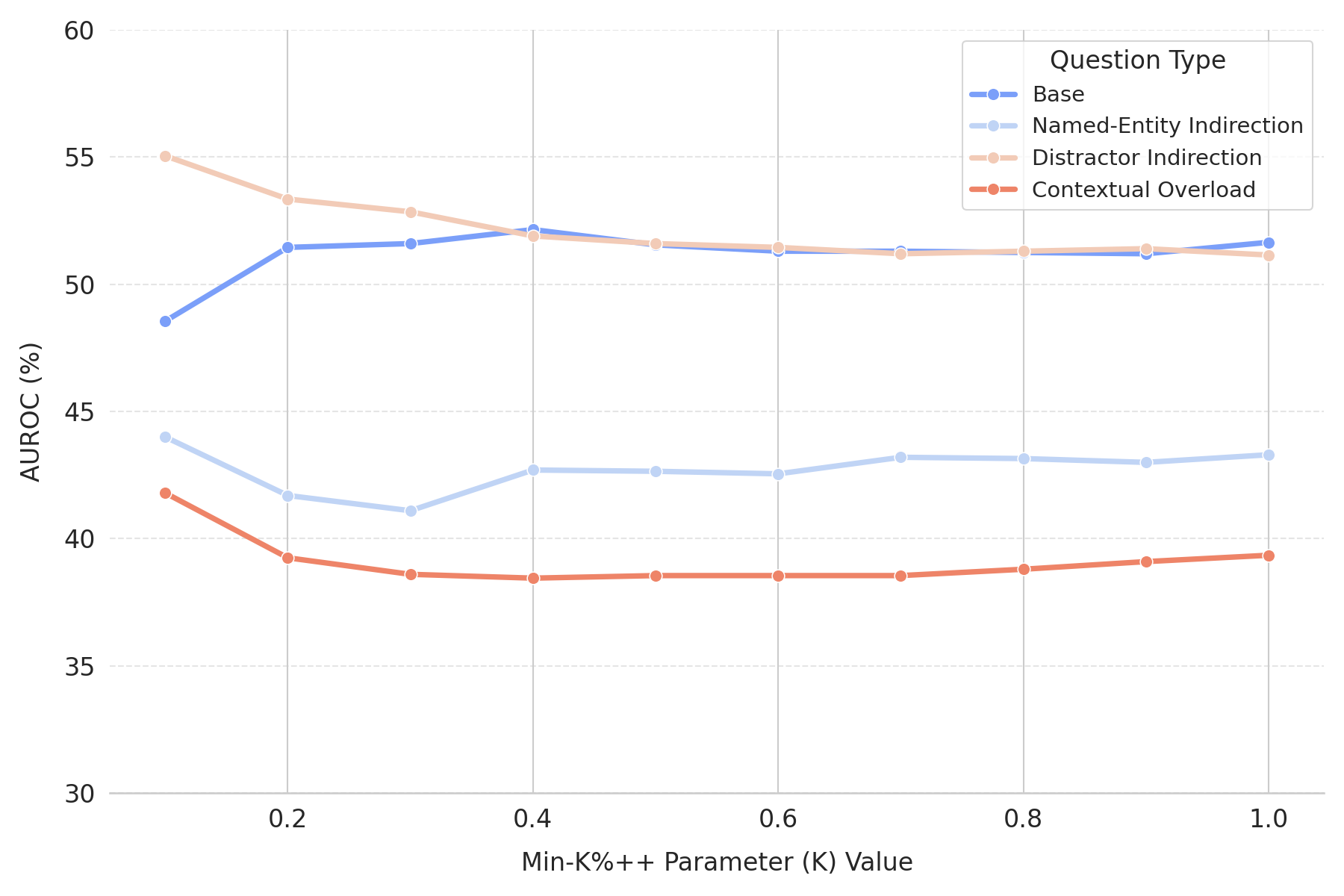}
    \caption{AUROC performance across Min-K\%++ parameter \(K\) on LLaMA~3.1~8B. Obfuscation (particularly DI and CO) reduces pre-training detectability.}
    \label{fig:mink++}
\end{figure}

\begin{figure}[ht]
    \centering
    \includegraphics[width=0.48\textwidth, height=6cm]{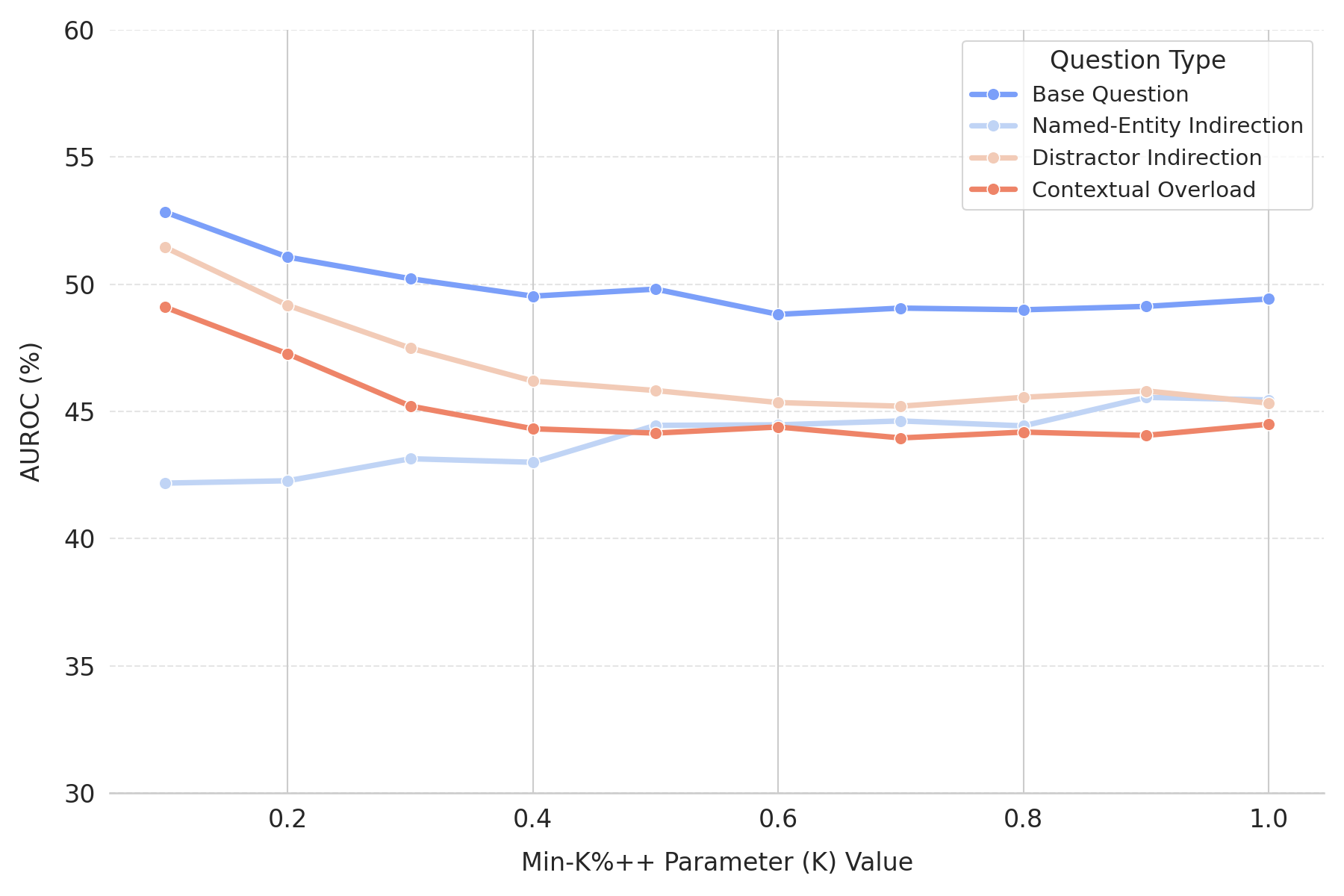}
    \caption{AUROC performance across Min-K\%++ parameter \(K\) on Mistral~7B~v0.1. Similar declines observed, confirming reduced memorization of obfuscated inputs.}
    \label{fig:mink++mistral}
\end{figure}
\subsection{Layer-wise Norm Drop Analysis}
\label{layerwise}
To explore how obfuscation impacts internal representations of a model, we compute the mean $\ell_2$ norm of each token’s hidden-state vector at every transformer layer (cf. Fig. \ref{fig:layernorm}). The $\ell_2$ norm $\lVert \mathbf{h}_{\ell,t}\rVert_2$ serves as a proxy for token activation strength or semantic richness. The mean layer-wise profile is calculated as:
\[
\bar{n}_{\ell} = \frac{1}{T}\sum_{t=1}^{T}\lVert \mathbf{h}_{\ell,t}\rVert_{2}
\]
A sharp drop in $\bar{n}_{\ell}$ signifies a \textit{semantic compression bottleneck}, where the network condenses distributed token representations into higher-level abstractions. 

For both LLaMA~3.1~8B and Mistral~7B~v0.1, base questions exhibit a relatively late norm drop (around Layer~14), whereas obfuscated inputs (NEI, DI, CO) show earlier compression, on average by \text{two layers sooner} (Layer~12). This \text{14-18\% earlier activation collapse} indicates that linguistic perturbations cause premature abstraction, reducing representational depth. In effect, models compress meaning too early (e.g., before resolving entity references or filtering distractors), leading to incomplete reasoning chains and lower output accuracy.

% \vspace{1mm}
% Overall, these intrinsic analyses reveal that obfuscation not only reduces output accuracy but also reshapes the internal dynamics of large language models—lowering self-confidence, disrupting recall alignment, and triggering early compression of semantic representations.

\vspace{-1mm}
\section{Conclusion}
\vspace{-2mm}
Our \texttt{ObfusQAte} shows novel light on LLMs robustness exhibited by LLMs' impairment caused by obfuscated versions of the base questions, hence opening a new avenue for interesting future work.
%Our results empirically suggest that while the models reliably handle base questions, the compounded challenges introduced by multiple layers of our \texttt{ObfusQAte} algorithm significantly impair performance, indicating that further research is required in this domain to robustly process and interpret such complex transformations in LLMs.

\section*{Limitations and Future Work}
While our current study focuses on a single QA dataset composed of factual questions in English, it provides a foundational step toward broader generalization. In future work, we aim to extend our dataset to include \textit{multilingual} data, particularly focusing on \textit{low-resource} languages to enhance inclusivity. To better capture the diversity of real-world QA tasks, we plan to incorporate additional categories such as mathematical reasoning, comprehension-based tasks, and translation challenges. Our future work will additionally implement obfuscation-based techniques in white-box settings to evaluate these systems in greater detail, promoting the development of equitable and highly robust AI systems.

\vspace{-5mm}

\section*{Ethics Statement}
All human annotators were compensated fairly and commensurate with their contributions to ensure that their time and efforts were respected and valued. The recruitment process followed ethical standards, and all participants provided informed consent regarding the use of their annotations. The paid models employed in our study were accessed strictly via valid subscriptions.
\vspace{-5mm}
\newpage
\section*{References}\label{sec:reference}
% \vspace{-10mm}

\bibliographystyle{lrec2026-natbib}
\bibliography{main}

% \vspace{-10mm}

% ------------------------------------------------------------------------
% COMMENTING OUT APPENDIX FOR LERC:
% -------------------------------------

\newpage
\appendix
\vspace{-20mm}
\section{Appendices}
\vspace{-20mm}

\subsection{ObfusQAte Prompts}
% \vspace{-22mm}
\label{sec:obfusprompts}

\label{A.1}
\begin{llmprompt} [Named-Entity Indirection]
\vspace{1mm}

1. Identify Named-Entities: 
First, find all the named entities in the question. Named entities can include people, organizations, locations, dates, etc.
\vspace{1mm}

2. Apply indirection to the named entities
    - Using synonyms for the named entities.\vspace{1mm}

    - Replacing named entities with pronouns.\vspace{1mm}

    - Using definitions or descriptions instead of the actual named entity.\vspace{1mm}

    - Introducing family relations, if applicable, or other indirect forms.
\vspace{1mm}

 3. Reframe the Question: 
 After transforming the named entities, rephrase the question so that it becomes more difficult to answer, while still maintaining its original meaning.
\vspace{1mm}

 4. Review the Final Question: 
 Ensure that the final question is significantly more challenging and requires more reasoning or external knowledge to answer correctly.

\vspace{2mm}

\end{llmprompt}
\vspace{-2mm}

\begin{llmprompt} [Distractor Indirection]
\vspace{2mm}

1. Identify Named Entities: First, find all the named entities in the question. Named entities can include people, organizations, locations, dates, etc.
\vspace{1mm}

2.Apply indirection to the named entities, this can involve:
\vspace{1mm}

- Using synonyms for the named entities.\vspace{1mm}

- Replacing named entities with pronouns.\vspace{1mm}

- Using definitions or descriptions instead of the actual named entity.\vspace{1mm}

- Introducing family relations, if applicable or apply other indirect forms.\vspace{1mm}

3. Add Very Little Context: Add a tiny bit of context (one short phrase or clause) to the question without giving away the answer.\vspace{1mm}

4. Add Distractor: Introduce distractors to make the question more challenging:\vspace{1mm}

 - Similar Concepts: Add semantically or topically related concepts that might confuse the answer (e.g. “Was Hamlet published the same year as Romeo and Juliet?”).\vspace{1mm}

 - Plausible Alternatives: Offer incorrect but reasonable alternatives that could distract the user (e.g. “Was Hamlet written by Marlowe, Shakespeare or Jonson?”).\vspace{1mm}

5. Rephrase the Question: 

Change the structure of the question so that it forces the user to think harder or search for answers in various places.
\vspace{1mm}

\end{llmprompt}

% \vspace{-30mm}

\begin{llmprompt} [Contextual Overload, before skip=5pt, after skip=5pt]
\vspace*{2mm}

1. Identify Named Entities: First, find all the named entities in the question. Named entities can include people, organizations, locations, dates, etc.
\vspace{2mm}

2. Apply indirection to the named entities, this can involve:
\vspace{2mm}

 - Using synonyms for the named entities.\vspace{1mm}

 - Replacing named entities with pronouns.\vspace{1mm}

 - Using definitions or descriptions instead of the actual named entity.\vspace{1mm}

 - Introducing family relations, if applicable, or apply other indirect forms.\vspace{2mm}
 
% \end{llmprompt}

% \begin{llmprompt} [Contextual Overload - continued, before skip=5pt, after skip=5pt]
% \vspace*{2mm}

3. Contextual Overload:
\vspace*{2mm}

- Irrelevant Details: Add information that is not directly relevant to the answer but might confuse or mislead the respondent.\vspace{2mm}

- Red Herring Facts: Include incorrect facts that might seem plausible.\vspace{2mm}

- Temporal/Spatial Context: Frame the question within a specific period or geographic location, requiring a more specific answer.
\vspace*{2mm}

4. Rephrase the Question: Change the structure of the question so that it forces the user to think harder or search for answers in various places.

\vspace*{1mm}
\end{llmprompt}

% \vspace{-30mm}
\vspace{-2em}

\subsubsection{Prompting strategies}
\label{sec:promptingstrategies}
\label{A.1.1}
\vspace{-3mm}

\begin{llmprompt} [Chain-of-Thought]
\vspace{1mm}

You are an expert at answering complex and obfuscated objective-type questions. 
\vspace{1mm}

Think step by step to deconstruct the question, identify the core information needed, and derive the correct answer.
\vspace*{1mm}

Finally, state the answer clearly and concisely. 
\vspace*{2mm}

Follow these steps:
\vspace*{1mm}

1. Read the entire question carefully, even if it includes extra indirection or distractors.
\vspace*{1mm}

2. Isolate the core query by stripping away any added layers of indirection, irrelevant details,or red herrings. 
\vspace*{1mm}

3. Reverse any transformations to recover the original meaning of the question. 
\vspace*{1mm}

4. Apply logical reasoning and your domain knowledge to determine the correct answer.
\vspace*{-2mm}

5. Finally, output only the concise final answer without showing any internal reasoning or extra text.
\vspace*{2mm}

\end{llmprompt}

\begin{llmprompt} [Few-shot Prompt]
\vspace*{2mm}

You are an assistant that answers only with the objective answer. Do not include any additional information.\\
\vspace*{1mm}

When responding, carefully review the examples that include both the base question and the modifications, and use these to infer the intended meaning of the asked question and deliver answer:
\vspace{2mm}

NAMED-ENTITY-INDIRECTION = """
\vspace{2mm}

Examples:\vspace{1mm}

Example 1:
base-Question: <INSERT YOUR EXAMPLE 1 HERE>
Answer: <GROUND TRUTH ANSWER 1>
\vspace{1mm}

Example 2:
base-Question: <INSERT YOUR EXAMPLE 2 HERE>
Answer: <GROUND TRUTH ANSWER 2>
"""
\vspace{2mm}

DISTRACTION INDIRECTION = """
\vspace{2mm}

Examples:\vspace{1mm}

Example 1:
base-Question:  <INSERT YOUR EXAMPLE 1 HERE>
Answer: <GROUND TRUTH ANSWER 1>
\vspace{1mm}

Example 2:
base-Question: <INSERT YOUR EXAMPLE 2 HERE>
Answer: <GROUND TRUTH ANSWER 2>
"""\vspace{2mm}

CONTEXTUAL-OVERLOAD = """
\vspace{2mm}

Examples:\vspace{1mm}

Example 1:
base-Question:  <INSERT YOUR EXAMPLE 1 HERE>
Answer: <GROUND TRUTH ANSWER 1>
\vspace{1mm}

Example 2:
base-Question: <INSERT YOUR EXAMPLE 2 HERE>
Answer: <GROUND TRUTH ANSWER 2>
"""\vspace{2mm}

\end{llmprompt}

% \vspace{-10mm}

\subsection{Annotation Workflow}

% \subsubsection{Annotator's Details}
\label{A.6}

\vspace{1em}
\noindent
The annotation process involves the following steps:
\begin{enumerate}
    \item \textbf{Synthetic Generation:} Each obfuscation question is produced by prompting the LLM with instructions to transform the base question into NEI, DI, and CO formats.
    \item \textbf{Human Review:} The annotators read the generated questions carefully and checked for:
        \emph{(i) Factual Consistency:} Does the obfuscated question still refer to the correct entity (i.e., capital of Australia) and avoid contradictory statements?
        
        \emph{(ii) Semantic Faithfulness:} Does the question still ask for the same information as the base question?
        
        \emph{(iii) Hallucinations:} Are there any introduced inaccuracies (e.g., attributing the capital city to the wrong country)?
    \item \textbf{Edits and Corrections:} The annotators make subtle wording adjustments to remove or fix any detected hallucinations while preserving the intended obfuscation style to prevent any semantic "drift" bias.
    \item \textbf{Final Verification:} Each revised question is confirmed to be semantically aligned with the original base query, ensuring that all four variants (Base, NEI, DI, CO) ask for the same underlying fact.
\end{enumerate}

Through this process, we obtain the final, human-verified version of the obfuscated questions, each pointing to the same \emph{ground truth} answer, \textbf{Canberra}. 

This example illustrates how our annotation workflow maintains consistency and factual accuracy even as the questions become increasingly complex.

% \subsubsection{ Inter-Annotator Reliability}
% To measure inter-annotator reliability, we compute \textit{Cohen’s Kappa} ($\kappa$) between the two annotators for entire the dataset across all the obfuscated variants. Cohen’s Kappa accounts for agreement occurring by chance, making it more robust than raw accuracy in evaluating categorical labeling tasks.

% We compute $\kappa$ as:

% \[
% \kappa = \frac{p_o - p_e}{1 - p_e}
% \]

% where $p_o$ is the observed agreement and $p_e$ is the expected agreement by chance.

% In our case, annotations were compared at the sentence level for semantic equivalence and factual alignment. The resulting Kappa score was $\kappa = 0.862$, indicating \textit{strong agreement} according to standard interpretation thresholds. Disagreements were resolved via adjudication to ensure high-quality supervision for evaluation.

% \vspace{2em}

\subsection{Qualitative Examples of Model Behavior}
\label{sec:reason}
\label{A.5}
\vspace{-1mm}
We present representative examples to illustrate model behavior under different obfuscation types. Correct (\textcolor{green}{green}) and incorrect (\textcolor{red}{red}) answers highlight successful reasoning and failure cases, respectively.
\vspace{5mm}

\vspace{-5mm}

\begin{tcolorbox}[colback=blue!5, colframe=blue!80!black, title={Example 1: Ground Truth --- \textbf{Anita Loos}}]
\vspace{1em}

\textbf{Base Question:} Who wrote the novel \textit{Gentlemen Prefer Blondes}?\\[1em]
\Answer{Claude 3.5 Sonnet Answer}{Anita Loos}{correct}
\Answer{Deepseek R1 Answer}{Anita Loos}{correct}
\Answer{GPT o3-mini Answer}{Anita Loos}{correct}
\Answer{GPT 4o Answer}{Anita Loos}{correct}

\textbf{NEI:} Who authored the widely read novel concerning the experiences of a blonde woman, a work that subsequently served as the inspiration for a renowned musical production, and whose author is also related, through her granddaughter, to the creator of a highly popular series of books chronicling the life of a young wizard?
\\[0.5em]

\vspace{1em}

\end{tcolorbox}

\begin{tcolorbox}[colback=blue!5, colframe=blue!80!black, title={Example 1: \textbf{Anita Loos} (continued)}]
\Answer{Claude 3.5 Sonnet Answer}{Lauren Weisberger}{incorrect}
\Answer{Deepseek R1 Answer}{Dodie Smith}{incorrect}
\Answer{GPT o3-mini Answer}{Gregory Maguire}{incorrect}
\Answer{GPT 4o Answer}{Anita Loos}{correct}

% \vspace{1em}
\textbf{DI:} Within the vibrant cultural landscape of the Jazz Age, which author, whose creative spark is said to have been ignited by anecdotes circulating within the opulent settings of Manhattan's grand hotels, wrote the satirical narrative focusing on the exploits of two captivating women known for their pursuit of wealth and advantageous marriages, a work frequently misattributed to the cousin of Anita Loos or a contemporary writer such as Elinor Glyn?
% \\[0.5em]

\Answer{Claude 3.5 Sonnet Answer}{(Anita Loos)}{correct}
\Answer{Deepseek R1 Answer}{Anita Loos}{correct}
\Answer{GPT o3-mini Answer}{Edith Wharton}{incorrect}
\Answer{GPT 4o Answer}{Edith Wharton}{incorrect}

% \end{tcolorbox}

% \begin{tcolorbox}[colback=blue!5, colframe=blue!80!black, title={Example 1: Ground Truth --- \textbf{Anita Loos}(continued)}]
% \vspace{1em}
\textbf{CO:} During the intellectually and artistically fertile period of the Roaring Twenties, characterized by the iconic flapper fashion and the infectious rhythms of jazz music, a certain comedic novel emerged, encapsulating the spirit of this dynamic era. 

% \end{tcolorbox}

 % \begin{tcolorbox}[colback=blue!5, colframe=blue!80!black, title={Example 1: \textbf{Anita Loos} (continued)}]

% \end{tcolorbox}

% \begin{tcolorbox}[colback=blue!5, colframe=blue!80!black, title={Example 1: \textbf{Anita Loos} (continued)}]

The granddaughter of the author celebrated for a compilation of poems frequently recited at Vassar College penned a narrative centered on the adventures of a blonde woman. This author, whose sibling was a distinguished academic specializing in ancient Greek literature, crafted a work that later achieved acclaim as a celebrated musical.
Could you identify the individual responsible for writing this narrative, a story depicting the escapades of a protagonist with fair hair, often perceived as naive, as she navigates the intricate social dynamics of a world populated by immense fortunes, aristocratic titles, and transatlantic voyages, while acknowledging that her contemporary, Anita Loos, was also a significant literary figure of that time?
% \vspace{1em}
\\[0.5em]
\Answer{Claude 3.5 Sonnet Answer}{Edith Wharton}{incorrect}
\Answer{Deepseek R1 Answer}{Lorelei Lee}{incorrect}
\Answer{GPT o3-mini Answer}{Anita Loos}{correct}
\Answer{GPT 4o Answer}{Evelyn Waugh}{incorrect}
\end{tcolorbox}

%---------------------------------------------
% Example 2
%---------------------------------------------
\begin{tcolorbox}[colback=blue!5, colframe=blue!80!black, title={Example 2: Ground Truth --- \textbf{Walter}}]

\textbf{Base Question:} What is Bruce Willis' real first name?
\Answer{Claude 3.5 Sonnet Answer}{Walter}{correct}
\Answer{Deepseek R1 Answer}{Walter}{correct}
\Answer{GPT o3-mini Answer}{Walter}{correct}
\Answer{GPT 4o Answer}{Walter}{correct}

\vspace{-1em}
\textbf{NEI:} The actor who played John McClane had one name and the other part of the name sounds similar to \texttt{Walter}. What is his first name?\\[0.5em]
\vspace{-1em}

\Answer{Claude 3.5 Sonnet Answer}{Walter}{correct}
\Answer{Deepseek R1 Answer}{Walter}{correct}
\Answer{GPT o3-mini Answer}{Bruce}{incorrect}
\Answer{GPT 4o Answer}{Walter}{correct}
\vspace{-0.5em}

\textbf{DI:} An actor renowned for portraying resilient, often wisecracking, characters in action films chose a stage name that resonated with a strong, assertive image. Before achieving global fame, this performer adopted a first name that diverges from the more common appellation he was given at birth. Knowing that his career took off in the 1980s and that he has German ancestry, what was his birth first name?
\vspace{-0.2em}
% \\[0.5em]
\Answer{Claude 3.5 Sonnet Answer}{Bruce}{incorrect}
\Answer{Deepseek R1 Answer}{Walter}{correct}
\Answer{GPT o3-mini Answer}{Bruce}{incorrect}
\Answer{GPT 4o Answer}{Bruce}{incorrect}
% \end{tcolorbox}

% \begin{tcolorbox}[colback=blue!5, colframe=blue!80!black, title={Example 2:  \textbf{Walter} (continued)}]
\vspace{-1em}
\textbf{CO:} While he adopted a stage name synonymous with Hollywood action stardom, the actor who brought to life iconic characters known for their grit and resilience entered the world bearing a different designation. This individual, whose surname suggests a determined disposition, hails from a family with roots extending to post-war Germany, a nation undergoing a period of significant cultural redefinition in the mid-20th century. Though he embraced a persona associated with American action heroes, his given name actually shares etymological roots with a term denoting a military commander. Was this birth name more commonly associated with artistic figures, biblical figures, or a character in World War I? Was it Arnold, Walter, or Bruno?
% \vspace{0.5em}

% \end{tcolorbox}

% \begin{tcolorbox}[colback=blue!5, colframe=blue!80!black, title={Example 2:  \textbf{Walter} (continued)}]

% \vspace{-1em}

% \\[0.2em]
\Answer{Claude 3.5 Sonnet Answer}{Sylvester}{incorrect}
\Answer{Deepseek R1 Answer}{Arnold}{incorrect}
\Answer{GPT o3-mini Answer}{Arnold}{incorrect}
\Answer{GPT 4o Answer}{Bruce}{incorrect}

\vspace{-1em}
\end{tcolorbox}

%---------------------------------------------
% Example 3
%---------------------------------------------
\begin{tcolorbox}[colback=blue!5, colframe=blue!80!black, title={Example 3: Ground Truth --- \textbf{United States / United States of America}}]

\textbf{Base Question:} In which country was Emilio Estevez born?\\[0.5em]
\Answer{Claude 3.5 Sonnet Answer}{United States}{correct}
\Answer{Deepseek R1 Answer}{United States}{correct}
\Answer{GPT o3-mini Answer}{United States}{correct}
\Answer{GPT 4o Answer}{United States}{correct}

% \vspace{1em}
\textbf{NEI:} In what nation, where the silver screen is celebrated and its sibling shares a patronym, did the offspring of the ``Apocalypse Now'' narrator first draw breath?\\[0.5em]

% \end{tcolorbox}

% \begin{tcolorbox}[colback=blue!5, colframe=blue!80!black, title={Example 3: \textbf{United States} (continued)}]

\Answer{Claude 3.5 Sonnet Answer}{Phillipines}{incorrect}
\Answer{Deepseek R1 Answer}{United States}{correct}
\Answer{GPT o3-mini Answer}{United States}{correct}
\Answer{GPT 4o Answer}{United States}{correct}

% \vspace{1em}
\textbf{DI:} Given the backdrop of his father's artistic journey, in what nation did Martin Sheen's eldest offspring first draw breath, noting that he shares a moniker with a saint?
% \vspace{1em}
\\[0.5em]
\Answer{Claude 3.5 Sonnet Answer}{Spain}{incorrect}
\Answer{Deepseek R1 Answer}{United States}{correct}
\Answer{GPT o3-mini Answer}{United States}{correct}
\Answer{GPT 4o Answer}{United States}{correct}

% \vspace{1em}

\textbf{CO:} In the nation renowned for its Hollywood heart and where a distinguished family, including the son of Martin Sheen (recognized by a title echoing a medieval military rank), first breathed life, despite his father's deep connections to another land celebrated for its shamrocks and folklore?

% \end{tcolorbox}

% \begin{tcolorbox}[colback=blue!5, colframe=blue!80!black, title={Example 3: Ground Truth --- \textbf{United States / United States of America}(continued)}]

% \vspace{1em}

% \vspace{1em}
% \\[0.5em]
\Answer{Claude 3.5 Sonnet Answer}{United States of America}{correct}
\Answer{Deepseek R1 Answer}{United States}{correct}
\Answer{GPT o3-mini Answer}{United States}{correct}
\Answer{GPT 4o Answer}{United States}{correct}

\end{tcolorbox}

\vspace{-3mm}

\subsection{Example Questions}
\label{sec:example}
\label{A.7}
\vspace{-2mm}

In this section, we present a collection of sample questions along with their corresponding answers produced by GPT-4o and Claude 3.5 Sonnet, the top-performing models on ObfusQA. We have predominantly selected questions that our scoring method has identified as correct. By presenting these curated examples, we try to showcase the adept answering abilities and current proficiency of these LLMs. (cf. Appendix: table: \ref{tab:gpt-claude-eg1}, \ref{tab:gpt-claude-eg2}, \ref{tab:gpt-claude-eg3}).
\vspace{2mm}

\begin{table*}[ht]

\centering
\caption{\textbf{Example 1:} Sample Queries and Model Answers from \textbf{GPT-4o} and \textbf{Claude 3.5 Sonnet}}
\label{tab:sample-questions}
\begin{tabularx}{\textwidth}{lX X}
\toprule
\textbf{Query Type} & \textbf{GPT-4o} & \textbf{Claude 3.5 Sonnet} \\
\midrule

%====================== Base Question ======================
\toprule
\textbf{Base Question} 
& 
\textit{Query:} What is the largest ocean on Earth?\newline
\textit{\textbf{Ground Truth:}} Pacific Ocean
& 
\textit{Query:} What is the chemical symbol for silver?\newline
\textit{\textbf{Ground Truth:}} Ag \\

%================ Named-Entity Indirection ================
\midrule
\textbf{Named-Entity Indirection} 
& 
\textit{Query:} Which body of water, known as the one bordering the Asian and American continents, and also referred to as the one Ferdinand Magellan crossed, holds the title for greatest surface area among all the world's interconnected hydrosphere components?\newline
\textit{Answer:} The Pacific Ocean
&
\textit{Query:} What alphanumeric designator is assigned to the element whose monetary applications historically rivaled those of aurum...?\newline
\textit{Answer:} Ag \\

%================ Distraction Indirection ==================
\midrule
\textbf{Distraction Indirection} 
& 
\textit{Query:} Considering its vastness and the ring of fire that surrounds it, which watery expanse, often navigated by vessels crossing from the land of the rising sun to the Americas, reigns supreme in terms of surface area when compared to the Atlantic, Indian, Arctic, and Southern oceans? \newline
\textit{Answer:} The Pacific Ocean
&
\textit{Query:} Considering its vastness and the ring of fire that surrounds it, which watery expanse, often navigated by vessels crossing from the land of the rising sun to the Americas, reigns supreme in terms of surface area when compared to the Atlantic, Indian, Arctic, and Southern oceans? \newline
\textit{Answer:} Ag \\
\midrule
%================ Contextual Overload ======================
\textbf{Contextual Overload} 
& 
\textit{Query:} During the administration of the 45th U.S. President, amidst debates about climate change and maritime boundaries, what body of water, often associated with tales of krakens and explorations by Magellan, holds the greatest surface area, exceeding that of the Atlantic and Indian combined, despite some cartographers disputing its northernmost reaches due to Arctic ice formations, mistakenly suggesting the Arctic Ocean's dominance in size? \newline
\textit{Answer:} The Pacific Ocean
&
\textit{Query:} During the administration of the 45th U.S. President, amidst debates about climate change and maritime boundaries, what body of water, often associated with tales of krakens and explorations by Magellan, holds the greatest surface area, exceeding that of the Atlantic and Indian combined, despite some cartographers disputing its northernmost reaches due to Arctic ice formations, mistakenly suggesting the Arctic Ocean's dominance in size? \newline
\textit{Answer:} Ag \\

\bottomrule
\end{tabularx}
\label{tab:gpt-claude-eg1}
\end{table*}

\begin{table*}[ht]
\centering
\caption{\textbf{Example 2:} Sample Queries and Model Answers from \textbf{GPT-4o} and \textbf{Claude 3.5 Sonnet}}
\label{tab:sample-questions}
\begin{tabularx} {\linewidth}{@{}p{3cm} X X@{}}
\toprule
\textbf{Query Type} & \textbf{GPT-4o} & \textbf{Claude 3.5 Sonnet} \\
\midrule

%====================== Base Question ======================
\toprule
\textbf{Base Question} 
& 
\textit{Query:} What is the smallest country in the world?\newline
\textit{\textbf{Ground Truth:}} Vatican City.
& 
\textit{Query:} Who painted the ceiling of the Sistine Chapel?\newline
\textit{\textbf{Ground Truth:}} Michelangelo \\

%================ Named-Entity Indirection ================
\midrule
\textbf{Named-Entity Indirection} 
& 
\textit{Query:} What is the designation given to the sovereign state with the smallest area, an entity whose leader governs from a location within the boundaries of the city known as the center of the Catholic Church, a religious institution led by the successor to Saint Peter?\newline
\textit{Answer:} Vatican City
&
\textit{Query:} Who was the artistic genius commissioned to adorn the Vatican's most sacred chamber, a space renowned for its depiction of biblical narratives across its uppermost surface?\newline
\textit{Answer:} Michelangelo \\

%================ Distraction Indirection ==================
\midrule
\textbf{Distraction Indirection} 
& 
\textit{Query:} Considering its unique relationship with the Holy See, which sovereign nation, possessing the least expanse of terrestrial territory when measured by conventional cartographical means, is globally recognized for its diminutive size, rivaling others such as Nauru or Tuvalu?\newline
\textit{Answer:} Vatican City
&
\textit{Query:} Considering his extensive Florentine background and influence from Donatello, who was commissioned by the same pontiff who also initiated the rebuilding of St. Peter's Basilica to decorate the palatial sacred space, and not Raphael, complete the frescoes?\newline
\textit{Answer:} Michelangelo \\

%================ Contextual Overload ======================
\midrule
\textbf{Contextual Overload} 
& 
\textit{Query:} Considering the spectrum of political geography, from sprawling continental federations to ephemeral assertions of territorial autonomy, and acknowledging the unique status of polities whose authority transcends mere physical boundaries: Which internationally acknowledged sovereign entity, whose terrestrial footprint is eclipsed even by the gardens of the aforementioned pontifical enclave, represents the nadir of territorial extension among independent nation-states absent ongoing disputes over legitimacy, and falls demonstrably below the minimum spatial threshold required for county-level designation within that transatlantic republic founded upon ideals of representative self-governance, thus distinguishing it from insular micro-polities adrift within the ocean named for peacefulness, whose census rolls scarcely register four-figure population counts while also explicitly ignoring a known sea platform from WW2?\newline
\textit{Answer:} Vatican City
&
\textit{Query:} Amidst the artistic fervor of the Renaissance, and considering the era's patronage system where familial influence often dictated commissions, who was the individual, renowned for sculpting David and whose artistic journey was significantly shaped by his Florentine origins, responsible for the fresco adorning the apex of the papal sanctuary inaugurated by Sixtus IV, a space where cardinals convene to elect the spiritual leader of a global faith, even though some falsely attribute sections depicting the Genesis narrative to Raphael due to their shared period of prolific creation and similar mastery of perspective?\newline
\textit{Answer:} Michelangelo Buonarroti \\

\bottomrule
\end{tabularx}
\label{tab:gpt-claude-eg2}
\end{table*}

\begin{table*}[ht]
\centering

\caption{\textbf{Example 3:} Sample Queries and Model Answers from \textbf{GPT-4o} and \textbf{Claude 3.5 Sonnet}}
\label{tab:sample-questions}
\begin{tabularx}{\textwidth}{lX X}
\toprule
\textbf{Query Type} & \textbf{GPT-4o} & \textbf{Claude 3.5 Sonnet} \\
\midrule

%====================== Base Question ======================
\toprule
\textbf{Base Question} 
& 
\textit{Query:} What is the name of the Earth's natural satellite?\newline
\textit{\textbf{Ground Truth:}} The Moon
& 
\textit{Query:} What is the name of the process by which plants make their own food?\newline
\textit{\textbf{Ground Truth:}} Photosynthesis \\
\midrule

%================ Named-Entity Indirection ================
\textbf{Named-Entity Indirection} 
& 
\textit{Query:} What is the appellation given to the celestial body, orbiting our pale blue dot, that ancient cultures revered as a deity of the night, and which Neil Armstrong visited?\newline
\textit{Answer:} The Moon
&
\textit{Query:} What is the appellation given to the metabolic pathway, essential for autotrophic sustenance, wherein verdant organisms, descendants of archaeplastida, synthesize organic compounds from ethereal carbon dioxide and aqueous solutions, utilizing radiant energy?\newline
\textit{Answer:} Photosynthesis \\
\midrule

%================ Distraction Indirection ==================
\textbf{Distraction Indirection} 
& 
\textit{Query:} Contemplating the celestial ballet of our solar system, and noting the distinct absence of rings around the third rock from the sun: What solitary companion, whose rhythmic dance governs oceanic tides and whose ethereal glow has fueled countless works of art and ancient myth, is also called "that thing in the sky" by a particular group that originated in a city named in a similar way to one that is in Egypt, distinctly separate satellite from the fourth rock of our system?\newline
\textit{Answer:} Moon, that white thing
&
\textit{Query:} Considering the vital role of greenery in sustaining ecosystems, what is the denomination attributed to the endogenous alimentary fabrication observed within verdant organisms, as opposed to, say, chemosynthesis or the metabolic processes of fungi?\newline
\textit{Answer:} Photosynthesis \\

%================ Contextual Overload ======================
\midrule
\textbf{Contextual Overload} 
& 
\textit{Query:} During the space race era, amidst whispers of Soviet advancements and American aspirations, what celestial body, often personified in folklore as a feminine entity and orbiting our terrestrial sphere, reflects sunlight back to the third planet from the sun, a planet sometimes referred to as "Gaia" by environmentalists, and is not Phobos or Deimos?\newline
\textit{Answer:} The Moon
&
\textit{Query:} During the reign of the Sun King in France, a process vital for sustaining life on Earth was being studied, though its intricacies were not yet fully understood. If one were to observe a verdant organism, a member of the botanical kingdom, engaging in a self-nourishing activity within the gardens of Versailles, what term would best describe this essential, energy-converting function, keeping in mind that contemporary scientists mistakenly believed soil was the plant's primary food source?\newline
\textit{Answer:} Photosynthesis \\

\bottomrule
\end{tabularx}
\label{tab:gpt-claude-eg3}
\end{table*}

\subsection{Hosting \& Maintenance}

The ObfusQAte dataset is publicly hosted on Hugging Face.
The repository provides structured dataset configurations, version control, and direct integration with the \texttt{datasets} library for seamless loading and evaluation. Hosting on Hugging Face ensures long-term accessibility, reproducibility, and transparent version tracking.

The dataset will be actively maintained by the authors. Any updates, corrections will be released through versioned commits in the same repository.

\subsection{Data Licensing}

ObfusQAte is released under the \textbf{MIT License}. This permissive license allows use, modification, distribution, and reproduction of the dataset for both academic and commercial purposes, provided that appropriate attribution is given to the original authors.

\subsection{Intended Usage}

The \textbf{ObfusQAte} framework and dataset are designed primarily for research and development purposes, with the goal of evaluating and improving large language models (LLMs) resilience to obfuscated or indirectly phrased queries. By systematically challenging these LLMs with varied levels of semantic, distractive, and contextual complexity, researchers can pinpoint vulnerabilities in current models and devise strategies (e.g., improved prompt conditioning, fine-tuning/alignment, or adversarial training) to enhance their factual consistency. Importantly, ObfusQAte also exposes instances where models rely on mere memorization of pre-trained data rather than genuine reasoning, thereby highlighting the need for approaches that foster true understanding. We discourage using this dataset to deliberately deceive or mislead end-users, as the intent is to foster robust, transparent AI systems that better serve real-world needs.

\end{document}